\documentclass[acmsmall,screen]{acmart}
\usepackage{soul}

\usepackage{amssymb}
\usepackage{makecell}
\usepackage{multirow}
\usepackage{enumitem}
\usepackage{array}
\usepackage{graphicx}
\usepackage{psfrag}
\usepackage{subfigure}
\usepackage{colortbl}
\usepackage{bbding}
\usepackage{wrapfig}
\usepackage{url}
\usepackage{supertabular}
\usepackage{adjustbox}
\usepackage{threeparttable}
\usepackage{tikz}
\usepackage{amsmath}
\usepackage{booktabs}
\usepackage{lscape}
\usepackage{verbatim}
\usepackage{longtable}
\usepackage{stfloats}
\usepackage[misc]{ifsym}
\usepackage{algorithm}
\usepackage{algpseudocode}
\usepackage{bm}
\usepackage{mathrsfs}
\usepackage[figuresright]{rotating}
\usepackage{lipsum}

\newtheorem{MyDef}{Definition}[subsection]

\AtBeginDocument{%
  \providecommand\BibTeX{{%
    \normalfont B\kern-0.5em{\scshape i\kern-0.25em b}\kern-0.8em\TeX}}}

\usepackage{xspace}
\makeatletter 
\DeclareRobustCommand\onedot{\futurelet\@let@token\@onedot}
\def\@onedot{\ifx\@let@token.\else.\null\fi\xspace}

\def\eg{\emph{e.g}\onedot} 

\def\ie{\emph{i.e}\onedot}

\def\etc{\emph{etc}\onedot}

\makeatother

\setcopyright{acmcopyright}

\acmJournal{CSUR}

\begin{document}

\title{Deep Learning for Cross-Domain Few-Shot Visual Recognition: A Survey}

\author{Huali Xu}
\affiliation{%
  \institution{College of Computer Science, Nankai University}
  \city{Tianjin}
  \country{China}
  \postcode{300071}
}
\affiliation{%
  \institution{CMVS, University of Oulu}
  \streetaddress{Pentti Kaiteran katu 1}
  \city{Oulu}
  \country{Finland}
  \postcode{90570}
}
\email{huali.xu@oulu.fi}

\author{Shuaifeng Zhi}  \authornote{Li Liu and Shuaifeng Zhi are corresponding authors.}
\affiliation{%
  \institution{College of Electronic Science and Technology, National University of Defense Technology}
  \city{Changsha}
  \state{Hunan}
  \country{China}}
\email{zhishuaifeng@outlook.com}

\author{Shuzhou Sun}
\affiliation{%
  \institution{CMVS, University of Oulu}
  \streetaddress{Pentti Kaiteran katu 1}
  \city{Oulu}
  \country{Finland}
  \postcode{90570}
}
\email{shuzhou.sun@oulu.fi}

\author{Vishal M. Patel}
\affiliation{%
  \institution{Johns Hopkins University}
  \streetaddress{3400 N. Charles Street}
  \city{Baltimore}
  \state{Maryland}
  \country{USA}
}
\email{vpatel36@jhu.edu}

\author{Li Liu}  
\authornotemark[1]
\affiliation{
  \institution{College of Electronic Science and Technology, National University of Defense Technology}
  \city{Changsha}
  \state{Hunan}
  \country{China}}
\email{liuli_nudt@nudt.edu.cn}

\renewcommand{\shortauthors}{Huali Xu, Shuaifeng Zhi, Shuzhou Sun, Vishal M. Patel and Li Liu.}

\begin{abstract}
  While deep learning excels in computer vision tasks with abundant labeled data, its performance diminishes significantly in scenarios with limited labeled samples. To address this, Few-shot learning (FSL) enables models to perform  the target tasks with very few labeled examples by leveraging prior knowledge from related tasks. However, traditional FSL assumes that both the related and target tasks come from the same domain, which is a restrictive assumption in many real-world scenarios where domain differences are common. To overcome this limitation, Cross-domain few-shot learning (CDFSL) has gained attention, as it allows source and target data to come from different domains and label spaces. This paper presents the first comprehensive review of Cross-domain Few-shot Learning (CDFSL), a field that has received less attention compared to traditional FSL due to its unique challenges. We aim to provide both a position paper and a tutorial for researchers, covering key problems, existing methods, and future research directions. The review begins with a formal definition of CDFSL, outlining its core challenges, followed by a systematic analysis of current approaches, organized under a clear taxonomy. Finally, we discuss promising future directions in terms of problem setups, applications, and theoretical advancements.
\end{abstract}





\newcommand\blfootnote[1]{%
\begingroup
\renewcommand\thefootnote{}\footnote{#1}%
\addtocounter{footnote}{-1}%
\endgroup
}
\blfootnote{This work was partially supported by National Key Research and Development Program of China (No. 2021YFB3100800), the Academy of Finland under grant (331883), Infotech Project FRAGES, the National Natural Science Foundation of China under Grant (61872379,  62022091,  62201603) and the Chinese Postdoctoral Science Foundation under Grant (2023TQ0088, GZC20233539).}

\maketitle


\section{Introduction}
Deep learning~\cite{dl} has been highly successful in computer vision~\cite{sg1,od1,app-detection,zhou2024diffdet4sar,li2024predicting,yang2024saratr,LiSARATRX25}, largely due to the availability of large-scale labeled datasets. However, in many practical scenarios, obtaining such large amounts of labeled data is difficult or costly. To address this challenge, Few-shot learning (FSL) aims to enable models to learn new tasks with only a limited number of labeled samples. Consequently, this problem has garnered significant attention in both academia and industry due to its broad real-world applications. While humans can easily distinguish between objects after seeing only a few examples, machines struggle to achieve similar efficiency. In domains such as natural scene images, large datasets are readily available, but FSL is crucial in scenarios where collecting large amounts of data is difficult. Since the problem was first introduced in 2006~\cite{fsl-1}, numerous methods have been proposed to tackle the challenges of FSL~\cite{fslsurvey,fslsurvey22,fslsurvey20,fsl18,fslsurvey1}.

\begin{figure}
	\centering
  \vspace{-0.3cm}
 	\includegraphics[width=0.9\linewidth]{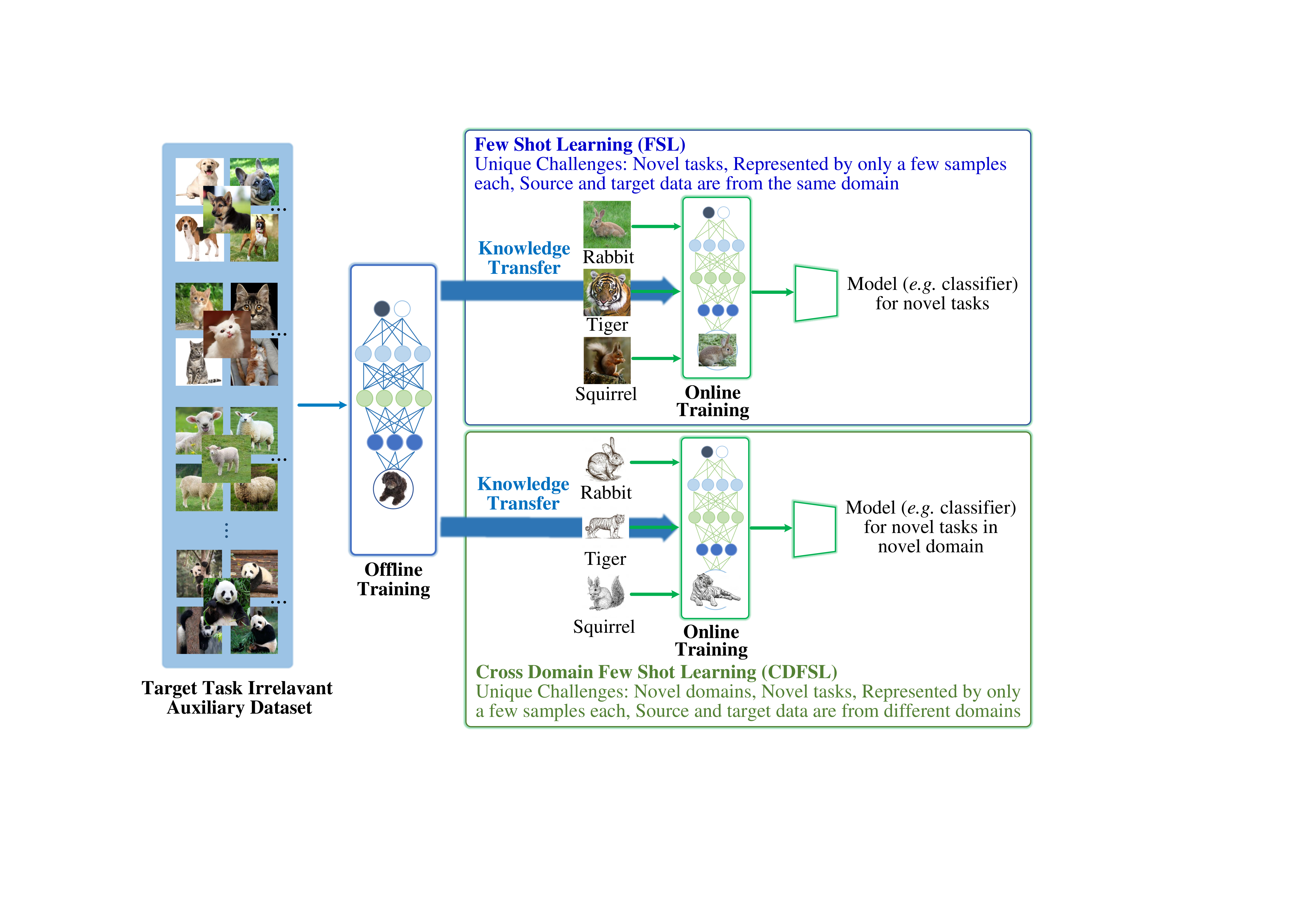}
  \vspace{-0.3cm}
	\caption{\textcolor{black}{The difference of few-shot learning and cross-domain few-shot learning.}}
 \vspace{-0.4cm}
	\label{int}
\end{figure}

\begin{figure}
	\centering
  \vspace{-0.3cm}
        \includegraphics[width=\linewidth]{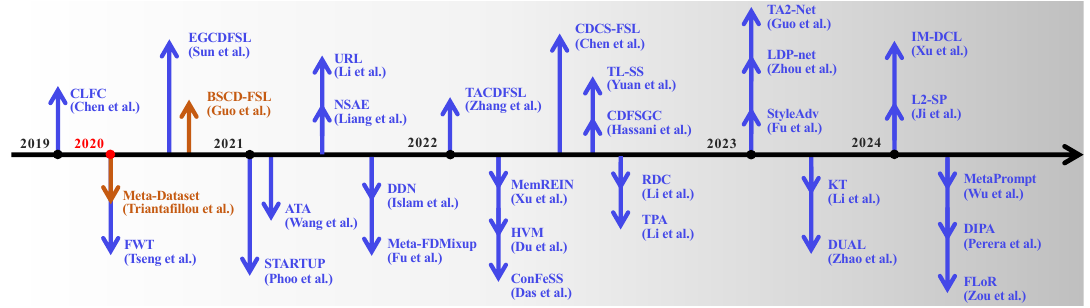}
 \vspace{-0.5cm}
	\caption{Chronological milestones of CDFSL from 2019 to the present, including representative CDFSL approaches and related benchmarks. Key events include the release of Meta-Dataset~\cite{meta-dataset} and BSCD-FSL~\cite{bscd-fsl} in 2020, the introduction of pioneering works such as~\cite{feature-wise}, and subsequent contributions like~\cite{feature_reweight_1,lscdfsl}. Later works~\cite{st,dynamic,hybrid_1,hybrid_4,hybrid_2} explored new setups, while~\cite{boosting,ata,data_target_1,feature_reweight_5,parameter_weight_2,confess,feature_reweight_9} focused on improving performance. Please see Section~\ref{methods} for details.}
 \vspace{-0.3cm}
	\label{imaging}
\end{figure}

With the development of FSL, challenges such as limited training data, domain variations, and task modifications have led to the emergence of various FSL variants, including semi-supervised FSL~\cite{semifsl}, unsupervised FSL~\cite{ufsl1,ufsl2}, zero-shot learning (ZSL)\cite{zsl1}, and cross-domain FSL (CDFSL)~\cite{feature-wise,bscd-fsl}, among others. These variants represent distinctive cases of FSL in terms of sample availability and domain learning. This paper focuses specifically on CDFSL variants. The traditional FSL problem assumes that both prior knowledge and target tasks come from the same domain, which is often restrictive in real-world applications. CDFSL addresses this issue by overcoming the domain gap between auxiliary data (which provides prior knowledge) and the target data in FSL tasks, as show in Figure~\ref{int}. For instance, in art image recognition tasks involving scribble, cartoon, or sketch images, FSL could theoretically leverage prior knowledge from related domains like cartoons and sketches. However, such data is often scarce due to copyright restrictions and the high cost of collection. As a result, researchers have turned to data-rich domains, such as natural scene images, to address the challenges of few-shot image recognition in the field of art.
However, the significant domain gap between these domains often leads to performance degradation in FSL. CDFSL faces challenges from both transfer learning and FSL, including domain gaps, class shifts, and the scarcity of labeled samples in the target domain, making it a more complex task. Since its formal introduction in 2020~\cite{feature-wise}, CDFSL has garnered widespread attention, with numerous methods published in top venues~\cite{bscd-fsl,st,dynamic,hybrid_1,feature_reweight_6}. Figure~\ref{imaging} presents the milestones of CDFSL technologies from 2019 to the present, showcasing representative CDFSL methods and related benchmarks.

So far, several surveys have provided comprehensive overviews and future directions for FSL~\cite{fsl18,fslsurvey,fslsurvey1,fslsurvey20,fslsurvey22}. For example,\cite{fsl18} categorizes FSL into experiential and conceptual learning, while\cite{fslsurvey} focuses on empirical risk minimization and defines FSL by experience, task, and performance, introducing CDFSL as a branch of FSL. Both~\cite{fslsurvey1} and~\cite{fslsurvey20} highlight CDFSL as a variant of FSL, discussing meta-learning approaches and benchmarks. Lastly,~\cite{fslsurvey22} offers a taxonomy based on prior knowledge and emphasizes that current methods have yet to fully tackle cross-domain challenges. Collectively, these works point to cross-domain learning as a promising area for future FSL research. Currently, there are two elementary surveys on CDFSL~\cite{wang2023survey,deng2023survey}. \cite{wang2023survey} classifies methods into benchmark, single source, and multiple source categories, while~\cite{deng2023survey} categorizes algorithms into data augmentation and feature alignment paradigms. In contrast, to stimulate future research and help newcomers better understand this challenging problem, this paper offers the first classification grounded in theoretical analysis and provides a comprehensive review, offering deeper insights into the core principles of CDFSL. Firstly, this paper compiles and analyzes a broad range of literature on the topic. An analysis of the reference index reveals that even before the formal introduction of CDFSL, some works had already tried to solve cross-domain issues within the FSL framework~\cite{clc, rffl}. Following its formal introduction as a branch of FSL, CDFSL has garnered significant attention. Additionally, we define CDFSL using both machine learning theory~\cite{ml,erm1} and transfer learning principles~\cite{tltheory}. Secondly, our analysis highlights that the unique challenge in CDFSL lies in the unreliable nature of two-stage empirical risk minimization. The details are discussed in Section~\ref{background}. To address these challenges, the paper organizes CDFSL research into four categories: $\mathcal{D}$-Extension, $\mathcal{H}$-Constraint, $\Delta$-Adaptation, and hybrid approaches. We also compile relevant datasets and benchmarks to evaluate the methods, and analyze their performance, as discussed in Sections~\ref{methods} and~\ref{performance}. Finally, we explore future research directions for CDFSL by considering three perspectives, including problem set-ups, applications, and theories, which provide a comprehensive understanding of the field and its potential for future growth. Contributions of this survey can be summarized as follows:

\begin{itemize}
    \item We analyzed existing CDFSL papers and provided a comprehensive survey. We also defined CDFSL formally, connecting it to classic ML~\cite{ml,erm1} and transfer learning theory~\cite{tltheory}. This helps guide future research in the field.
    \item We listed relevant learning problems for CDFSL with examples, clarifying their relation and differences. This helps position CDFSL among various learning problems. We also analyzed unique issues and challenges of CDFSL, helping to explore a scientific taxonomy for CDFSL work.
    \item We conducted an extensive literature review, organizing it into a unified taxonomy based on $\mathcal{D}$-Extension, $\mathcal{H}$-Constraint, $\Delta$-Adaptation, and hybrid approaches. We introduced applicable scenarios for each taxonomy to help discuss its pros and cons. We also presented datasets and benchmarks for CDFSL, summarizing insights from performance results to improve understanding of CDFSL methods.
    \item We proposed promising future directions for CDFSL in problem set-ups, applications, and theories, based on current weaknesses and potential improvements.
\end{itemize}

The remainder of this survey is organized as follows: Section \ref{background} provides an overview of CDFSL, including its definition, challenges, and taxonomy. Section \ref{methods} covers approaches to CDFSL in detail, while Section \ref{performance} presents performance results and evaluates methods. Section \ref{future} explores future directions in set-ups, applications, and theories. Finally, Section \ref{conclusion} concludes the survey.
\section{Background} \label{background}
In this section, we introduce key concepts related to CDFSL in Section \ref{key}, followed by formal definitions of supervised learning, FSL, domain adaptation (DA), and CDFSL with examples in Section \ref{definition}. Section \ref{related} discusses the connections and differences between CDFSL and related problems. In Section \ref{unique}, we cover the challenges that make CDFSL difficult. Finally, Section \ref{taxonomy} presents a unified taxonomy based on how existing works address these challenges.

\subsection{\textcolor{black}{Key Concepts}} \label{key}
To formally define the CDFSL problem, we begin by considering two key concepts: \emph{domain} and \emph{task}~\cite{tfsurvey,tf_2020}. These terms are often used inconsistently in the community, but a clear distinction between them helps in studying different transfer learning subproblems. In this paper, we follow the definitions provided by Pan and Yang's surveys\cite{tfsurvey,tf_2020,tfsurvey_2020}.

\begin{MyDef}
\label{def:Domain}
\textbf{Domain.} Given a feature space $\mathcal{X}$ and a marginal probability distribution $P(X)$,  where each input instance $\emph{\textbf{x}}\in\mathcal{X}$, a domain $\mathcal{D}=\{\mathcal{X}, P(X)\}$ consists of $\mathcal{X}$ and $P(X)$.
\end{MyDef}

In practice, a domain is often observed by several labeled or unlabeled data samples $X=\{\emph{\textbf{x}}\}_{i=1}^{N}$, where $N$ indicates the number of instances. For instance, if our learning task is image classification, and each input image is represented as a feature vector $\textbf{\emph{x}}$, \eg, by a deep convolution neural network (DCNN), then $\mathcal{X}$ is the space underlying the extracted feature vector. In general,  two different domains can differ in the feature space or the marginal distribution.

\begin{MyDef}
\label{def:Task}
\textbf{Task.}  Given a domain $\mathcal{D} = \{\mathcal{X}, P(X)\}$, a task  is composed of two components: a label space $\mathcal{Y} $ and a decision  function $f(\cdot)$ mapping each input sample to its belonging label, and is denoted as $\mathcal{T}=\{\mathcal{Y},f(\cdot)\}$. 
\end{MyDef}

Specifically, the $\textbf{\emph{x}}$ and $y$ represent the input data and supervision target. For a classification task $\mathcal{T}$, all labels $\textit{Y}^{\mathcal{T}}=\{y^{\mathcal{T}}_{1}, y^{\mathcal{T}}_{2}, ..., y^{\mathcal{T}}_{m}\} \in \mathcal{Y}$ are in the label space $\mathcal{Y}$, and $f(\cdot)$ can be learned from the training data $\textit{D}$=$\{\textbf{\emph{x}}_{i}, y_{i}\}^{N}_{i=1}$, where $\textbf{\emph{x}}_{i} \in \textit{X}$ and $y_{i} \in \textit{Y}$. From a probabilistic viewpoint, $f(\cdot)$ can be illustrated as a conditional probability distribution \textit{P(Y|X)}.

Comparatively speaking, for a learning problem, the domain describes the feature space $\mathcal{X}$ and the marginal distribution $P(X)$, while the task describes the output space $\mathcal{Y}$ and the conditional distribution $P(Y|X)$.

\subsection{\textcolor{black}{Problem Definition}} \label{definition}
In this subsection, we begin by defining vanilla supervised learning, followed by the definitions of FSL and domain adaptation (DA). We then introduce the definition of CDFSL, considering it a subproblem of both FSL and DA.

\vspace{-0.5cm}
\textcolor{black}{
\begin{MyDef}
\label{def:SupLearn}
\textbf{Vanilla Supervised Learning~\cite{erm1,erm2}.} Given a domain $\mathcal{D}$, consider a supervised learning task $\mathcal{T}$, a training set $\textit{D}^{train}$, and a test set $\textit{D}^{test}$. The goal of vanilla supervised learning is to learn a function $f(\cdot)$ for $\mathcal{T}$ on $\textit{D}^{train}$, such that $f(\cdot)$ performs well on $\textit{D}^{test}$, where $\{\textit{D}^{train}, \textit{D}^{test}\} \subseteq \mathcal{D}$.
\end{MyDef}
}

\begin{figure}[t]
	\centering
	 \vspace{-0.3cm}
        \includegraphics[width=\linewidth]{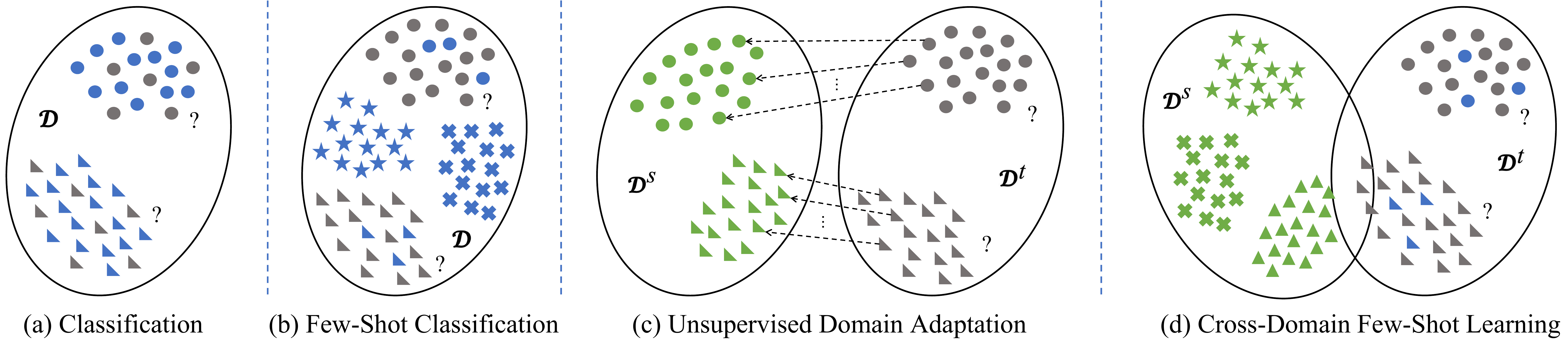}
  \vspace{-0.5cm}
	\caption{(a) the standard classification, (b) few-shot classification, (c) unsupervised domain adaptation, and (d) cross-domain few-shot classification. The different shapes represent different categories. $\mathcal{D}$ means domain, $\mathcal{D}^{s}$ and $\mathcal{D}^{t}$ specifically represent the source and target domains, respectively. Green and blue illustrate the source and target data. Gray represents the unlabeled test data, and `?' indicates predicting the test data. Dotted arrows indicate the adaptation process.}
 \vspace{-0.5cm}
	\label{dtfc}
\end{figure}

For example, an image classification task involves categorizing test set images $\textit{D}^{test}$ into classes using a model trained on $\textit{D}^{train}$. In classic classification, $\textit{D}^{train}$ has sufficient samples per class, like ImageNet~\cite{imagenet1} with 1000 classes and over 1000 samples per class. Note that $\textit{D}$ refers to the dataset, not the domain $\mathcal{D}$. Figure \ref{dtfc} (a) illustrates a standard supervised classification problem.

Like in vanilla supervised learning, the goal of FSL~\cite{fslsurvey} is to learn a model from the training set $\textit{D}^{train}$ for testing new samples. However, the key difference is that $\textit{D}^{train}$ in FSL contains very limited supervised data, making it challenging. Due to the few samples, many standard algorithms fail, often due to overfitting. To address this, prior knowledge is introduced from an auxiliary task $\mathcal{T}^{s}$ (source task). Typically, the label sets of source task $\mathcal{T}^{s}$ and target task $\mathcal{T}^{t}$ are disjoint ($\mathcal{Y}^{s} \cap \mathcal{Y}^{t}=\varnothing$). A formal definition of FSL is given below.

\begin{MyDef}
\label{def:FSL}
\textbf{Few Shot Learning (FSL)}~\cite{tfsurvey,fslsurvey}. Given a domain $\mathcal{D}$, a task $\mathcal{T}^{t}$ described by a \textit{T}-specific data set $\textit{D}^{t}$ with only a few supervised information available, and the task(s) $\mathcal{T}^{s}$ described by \textit{T}-irrelevant auxiliary data set(s) $\textit{D}^{s}$ with sufficient supervised information, FSL aims to learn a function $f(\cdot)$ for $\mathcal{T}^{t}$ by utilizing the limited supervised information in $\textit{D}^{t}$ and the prior knowledge in $(\mathcal{T}^{s}, \textit{D}^{s})$, where $\{\textit{D}^{s}, \textit{D}^{t}\} \subseteq \mathcal{D}$, $\textit{D}^{s} \cap \textit{D}^{t}=\varnothing$, and $\mathcal{T}^{s} \ne \mathcal{T}^{t}$.
\end{MyDef}

Specifically, take a few-shot classification task $\mathcal{T}^{t}$ as an example, we use the corresponding few-shot data pairs $\{(\textit{x}_{i}, \textit{y}_{i})\}^{N^{t}}_{i=1}$ to represent the input data and supervision target. In addition, $\mathcal{T}^{s}$ and $\{(\textit{x}_{i}, \textit{y}_{i})\}^{N^{s}}_{i=1}$ are utilized to indicate the conventional classification task and auxiliary data pairs, where $N^s \gg N^t$. $\mathcal{T}^{t}$ follows a \textit{C-way K-shot} training principle (\textit{C} indicates the number of classes, \textit{K} represents the sample numbers in each class). We learn a function $f$($\cdot$) for $\mathcal{T}^{t}$ from $\textit{D}^{t}$ and $(\mathcal{T}^{s}, \textit{D}^{s})$. Figure \ref{dtfc} (b) shows the few-shot classification (FSC) problem.

Beyond the FSL challenge, domain adaptation (DA)~\cite{dasurvey,dasurvey1,dasurvey2} is another key aspect of vanilla supervised learning. Like vanilla supervised learning, DA trains a model from $\textit{D}^{train}$. However, unlike vanilla supervised learning, the training and test sets in DA come from two different domains $\mathcal{D}^{s}$ and $\mathcal{D}^{t}$, \ie $\mathcal{D}^{s} \ne \mathcal{D}^{t}$. This violates the assumption in vanilla supervised learning that data must be independently and identically distributed. A formal definition of DA is provided below.

\begin{MyDef}
\label{def:DA}
\textbf{Domain Adaptation (DA)}~\cite{farahani2021brief,dasurvey1}. Given multiple different domains $\mathcal{D}=\left\{\mathcal{D}_{i}\right\}$ ($1 \leq i \leq I$, where $I$ denotes the total number of domains), which include source domains $\mathcal{D}^{s}$ associated with corresponding learning tasks $\mathcal{T}^{s}$, and target domains $\mathcal{D}^{t}$ with their learning tasks $\mathcal{T}^{t}$, where $\mathcal{D}=\left\{\mathcal{D}^{s}, \mathcal{D}^{t}\right\}$ and $\mathcal{D}^{s} \cap \mathcal{D}^{t} = \varnothing$. The goal of DA is to learn a target predictive function $f(\cdot)$ on $\mathcal{D}^{t}$ using prior knowledge from $(\mathcal{T}^{s}, \mathcal{D}^{s})$, where $\mathcal{D}^{s} \ne \mathcal{D}^{t}$, and $\mathcal{T}^{s}$ and $\mathcal{T}^{t}$ share the same label space.
\end{MyDef}

Supervised domain adaptation has been widely studied~\cite{dasurvey,dasurvey1}, so we use classification tasks to explain the unsupervised domain adaptation (UDA) problem~\cite{dasurvey}, where the target domain samples are unlabeled, as shown in Figure~\ref{dtfc} (c). Specifically, the target task $\mathcal{T}^{t}$ and its unlabeled data $\left\{\textit{x}_{i}\right\}^{N^{t}}_{i=1} \in \mathcal{D}^{t}$ are supported by the source task $\mathcal{T}^{s}$ with labeled data pair $\left\{(\textit{x}_{i}, \textit{y}_{i})\right\}^{N^{s}}_{i=1} \in \mathcal{D}^{s}$. Here, $\mathcal{D}^{s} \ne \mathcal{D}^{t}$, but $\mathcal{T}^{s}$ and $\mathcal{T}^{t}$ share the same label space. The goal is to learn a function $f(\cdot)$ for $\mathcal{T}^{t}$ by leveraging the data from both $\textit{D}^{t}$ and $(\mathcal{T}^{s}, \textit{D}^{s})$. Figure \ref{dtfc} (c) illustrates the unsupervised domain adaptation (UDA) classification problem.

Combining the challenges of FSL and DA, CDFSL~\cite{feature-wise,parameter_weight_4} predicts new samples using a model trained on $\{(\textit{x}_{i}, \textit{y}_{i})\}^{N^{t}}_{i=1}$ with prior knowledge from $\{(\textit{x}_{i}, \textit{y}_{i})\}^{N^{s}}_{i=1}$, where $N^{s} \gg N^{t}$. In CDFSL, the data $\{(\textit{x}_{i}, \textit{y}_{i})\}^{N^{s}}_{i=1}$ and $\{(\textit{x}_{i}, \textit{y}_{i})\}^{N^{t}}_{i=1}$ are drawn from different domains, $\mathcal{D}^{s}$ and $\mathcal{D}^{t}$ respectively, and they do not share the same label space, meaning $\mathcal{D}^{s} \ne \mathcal{D}^{t}$ and $\mathcal{T}^{s} \ne \mathcal{T}^{t}$. Compared to the FSL problem, where the data should be independent and identically distributed (i.i.d.), and the DA problem, which requires that tasks share the same label space, CDFSL breaks these constraints. Therefore, CDFSL inherits the challenges of both FSL and DA, making it a more challenging problem.
A definition of CDFSL is formally given below.

\begin{MyDef}
\label{def:CDFSL}
\textbf{Cross-Domain Few-Shot Learning (CDFSL).} Considering multiple different domains $\mathcal{D}=\left\{\mathcal{D}_{i}\right\}$ (1 $\leq$ i $\leq$ I, where I means the number of domains), including source domains $\mathcal{D}^{s}$ with sufficient information, along with corresponding learning tasks $\mathcal{T}^{s}$, and the target domains $\mathcal{D}^{t}$ with limited supervised information and FSL tasks $\mathcal{T}^{t}$, where $\mathcal{D}=\left\{\mathcal{D}^{s}, \mathcal{D}^{t}\right\}$ and $\mathcal{D}^{s} \cap \mathcal{D}^{t} = \varnothing$.
The goal of CDFSL is to learn a target predictive function $f_T(\cdot)$ on $\mathcal{D}^{t}$ with the help of prior knowledge from $(\mathcal{D}^{s}, \mathcal{T}^{s})$, where $\mathcal{D}^{s} \ne \mathcal{D}^{t}$, and $\mathcal{T}^{s} \ne \mathcal{T}^{t}$.
\end{MyDef}

In a cross-domain few-shot classification (CDFSC) problem~\cite{data_target_1,data_multi_1,data_multi_2}, as shown in Figure \ref{dtfc} (d), we similarly denote a source and a target classification task by $\mathcal{T}^{s}$ and $\mathcal{T}^{t}$, respectively. They are described by the data pairs $\{(\boldsymbol{x}_i^s,y^s_i)\}_{i=1}^{N^s} \subseteq \mathcal{D}^s$ and $\{(\boldsymbol{x}_i^t,y^t_i)\}_{i=1}^{N^t} \subseteq \mathcal{D}^{t}$, where $N^s\gg N^t$, $y^s_i\in\mathcal{Y}^s$, $y^t_i\in\mathcal{Y}^t$, $\mathcal{Y}^t\bigcap\mathcal{Y}^s=\varnothing$ (\ie, the source and target domains do not share the label space). Note that $\mathcal{D}^t$ and $\mathcal{D}^s$ are sampled from two different probability distributions $p$ and $q$, respectively, where $p \ne q$. The objective of the CDFSC is learning a classifier $f_{T}$($\cdot$) for $\mathcal{T}^{t}$ using $\mathcal{D}^{t}$ and $(\mathcal{T}^{s}, \mathcal{D}^{s})$.

CDFSL can be grouped into three categories based on why the image distributions differ: Fine-grained CDFSL ($FG$)~\cite{feature-wise,hybrid_7}, Art-based CDFSL ($Art$)~\cite{meta-dataset}, and Imaging way-based CDFSL ($IW$)~\cite{bscd-fsl,st,dynamic}. $FG$ refers to fine-grained class differences between $\mathcal{D}^s$ and $\mathcal{D}^t$, where $\mathcal{D}^t$ contains subclasses of $\mathcal{D}^s$. $Art$ involves differences in artistic styles like sketches, stick figures, and paintings. $IW$ deals with differences in imaging modalities, such as natural images in $\mathcal{D}^s$ and medical X-rays in $\mathcal{D}^t$, making it the most challenging of the three.

\subsection{\textcolor{black}{Closely Related Problems}} \label{related}
Since CDFSL is a sub-problem of both FSL and DA, in this section, we primarily focus on the connections and distinctions between CDFSL and the other sub-problems within FSL and DA, as illustrated in Table~\ref{rela}.
\begin{table}
\footnotesize
\centering
\caption{The connections and distinctions between the core related problems and CDFSL. $\mathcal{D}^{s}$ and $\mathcal{D}^{t}$ mean the source and target domain, respectively. And $\mathcal{Y}^{s}$ and $\mathcal{Y}^{t}$ represent the label space in the source and target tasks.}
\vspace{-0.2cm}
\setlength{\tabcolsep}{1.3mm}{
\begin{tabular}{lcccc}
\hline
\textbf{Problem} & \textbf{$\mathcal{D}^{s} \ne \mathcal{D}^{t}$} & \textbf{$\mathcal{Y}^{s} \ne \mathcal{Y}^{t}$} & \textbf{Limited $\mathcal{D}^{t}$} & \textbf{\textbf{Labeled data in $\mathcal{D}^{t}$}}       \\ 
 \hline
Semi-supervised domain adaptation (Semi-DA)~\cite{dasurvey1} & \Checkmark & \XSolidBrush & \XSolidBrush &  \Checkmark  \\
Unsupervised domain adaptation (UDA)~\cite{dasurvey,officehome} & \Checkmark & \XSolidBrush & \XSolidBrush & \XSolidBrush  \\
Universal domain adaptation (UniDA)~\cite{you2019universal,saito2020universal} & \Checkmark & \Checkmark & \XSolidBrush & \XSolidBrush  \\
Domain generalization (DG)~\cite{generalization_theory,dg11} & \Checkmark & \XSolidBrush & \XSolidBrush & Unseen $\mathcal{D}^{t}$  \\
Multi-task learning (MTL)~\cite{mtl1} & \XSolidBrush & \Checkmark & \XSolidBrush & \Checkmark   \\
Few-shot learning (FSL)~\cite{fslsurvey} & \XSolidBrush & \Checkmark & \Checkmark & \Checkmark   \\
Domain adaptation few-shot learning (DAFSL)~\cite{dafsl1,feature_reweight_8} & \Checkmark & \XSolidBrush & \Checkmark & \Checkmark   \\   \rowcolor{gray!30}
\textbf{Cross-domain few-shot learning (CDFSL)} & \Checkmark & \Checkmark & \Checkmark & \Checkmark   \\   
\hline
\end{tabular}}
\vspace{-0.5cm}
\label{rela}
\end{table}

\textit{Semi-supervised Domain Adaptation (Semi-DA)~\cite{dasurvey1}.} Semi-DA uses a large amount of supervised data in $\mathcal{D}^{s}$, along with a few labeled and many unlabeled samples in $\mathcal{D}^{t}$, to improve the performance of $\mathcal{T}^{t}$, with $\mathcal{D}^{s} \ne \mathcal{D}^{t}$ but the same label space. Similarly, CDFSL~\cite{feature-wise} also uses large supervised data in $\mathcal{D}^{s}$ and limited labeled data in $\mathcal{D}^{t}$, but without using many unsupervised samples from the target domain. Additionally, CDFSL differs in that the label spaces of $\mathcal{D}^{s}$ and $\mathcal{D}^{t}$ are different.

\textit{Unsupervised Domain Adaptation (UDA)~\cite{dasurvey,officehome}.} UDA utilizes a large amount of supervised data in $\mathcal{D}^{s}$ and a large amount of unlabeled data in $\mathcal{D}^{t}$ to improve the performance of $\mathcal{T}^{t}$. The distributions between $\mathcal{D}^{s}$ and $\mathcal{D}^{t}$ are different but related, \ie, $\mathcal{D}^{s} \ne \mathcal{D}^{t}$. And they share the same learning tasks. Like UDA, CDFSL~\cite{feature-wise} also uses a large amount of supervised data in $\mathcal{D}^{s}$ to improve the performance of $\mathcal{T}^{t}$ in $\mathcal{D}^{t}$, $\mathcal{D}^{s} \ne \mathcal{D}^{t}$. However, $\mathcal{D}^{t}$ in CDFSL has only a few amounts of supervised data, and the tasks of $\mathcal{D}^{s}$ and $\mathcal{D}^{t}$ are different.

\textit{Universal Domain Adaptation (UniDA)~\cite{you2019universal,saito2020universal}.} In UniDA, the source and target domains are different ($\mathcal{D}^{s} \ne \mathcal{D}^{t}$) and there is no prior knowledge of the label sets, meaning the source label set $\mathcal{Y}^{s}$ and target label set $\mathcal{Y}^{t}$ may overlap but also contain their own unique labels, creating a category gap. Specifically, $\mathcal{Y}^{s} \cap \mathcal{Y}^{t} \neq \varnothing$ and $\mathcal{Y}^{s} \setminus \mathcal{Y}^{t} \neq \mathcal{Y}^{t} \setminus \mathcal{Y}^{s} \neq \varnothing$. UniDA requires a model to either correctly classify the target sample if it belongs to the common label set $\mathcal{Y}^{s} \cap \mathcal{Y}^{t}$, or mark it as "unknown" if it does not. Similarly, in CDFSL~\cite{feature-wise}, the tasks in $\mathcal{D}^{s}$ and $\mathcal{D}^{s}$ differ, but CDFSL involves a few labeled target samples, and $\mathcal{Y}^{s} \cap \mathcal{Y}^{t} = \varnothing$. Additionally, there is no `unknown' class in CDFSL.

\textit{Domain Generalization (DG)~\cite{generalization_theory,dg11}.} DG uses a large amount of supervised data in $\textit{M}$ source domains $\mathcal{D}^{s}=\{\mathcal{D}^{s}_{i}|i=1,...,M\}$ to improve the performance on the unseen $\mathcal{D}^{t}$. The distributions of $\mathcal{D}^{s}$ and $\mathcal{D}^{t}$ are different but related, \ie $\mathcal{D}^{s} \ne \mathcal{D}^{t}$, and the tasks between $\mathcal{D}^{s}$ and $\mathcal{D}^{t}$ are same. Similar to DG, CDFSL~\cite{feature-wise} also uses a large amount of supervised data in $\mathcal{D}^{s}$. However, CDFSL is designed to perform well on the special $\mathcal{D}^{t}$ but not all unseen $\mathcal{D}^{t}$. Furthermore, the tasks of $\mathcal{D}^{s}$ and $\mathcal{D}^{t}$ are different in CDFSL, \ie $\mathcal{T}^{s} \ne \mathcal{T}^{t}$.

\textit{Domain Adaptation Few-shot Learning (DAFSL)~\cite{dafsl1,feature_reweight_8}.} DAFSL leverages a significant amount of supervised data in the source domain $\mathcal{D}^{s}$ and a limited number of labeled data in the target domain $\mathcal{D}^{t}$ to enhance the performance on $\mathcal{D}^{t}$. Although $\mathcal{D}^{s} \ne \mathcal{D}^{t}$, the learning tasks remain the same. Similarly, CDFSL~\cite{feature-wise} utilizes the same data configurations in both domains to train the function for task $\mathcal{T}^{t}$. However, in contrast to DAFSL, the learning tasks in $\mathcal{D}^{s}$ and $\mathcal{D}^{t}$ differ in CDFSL.

\textit{Multi-task Learning (MTL)~\cite{mtl1}.} MTL utilizes $M$ tasks from $\mathcal{D}$ to improve the performance of every task $\mathcal{T}_i$ (0 < i $\leq$ $M$). All $\{\mathcal{T}_i\}^M_{i=1}$ are different but related. Different from MTL, the data of $\mathcal{T}^{s}$ and $\mathcal{T}^{t}$ in CDFSL is from different domains $\mathcal{D}^{s}$ and $\mathcal{D}^{t}$, \ie $\mathcal{D}^{s} \ne \mathcal{D}^{t}$ and $\mathcal{T}^{s} \ne \mathcal{T}^{t}$, and the supervised data in $\mathcal{D}^{t}$ is limited.

\subsection{\textcolor{black}{Unique Issue and Challenge}} \label{unique}
In machine learning, prediction errors are typically unavoidable, making it impossible to achieve perfect predictions~\cite{ml}, \ie, the problem of unreliable empirical risk minimization (ERM). In this section, we begin by explaining the concept of empirical risk minimization (ERM)~\cite{erm1,erm2}. Next, we investigate the two-stage empirical risk minimization problem (TSERM)~\cite{tltheory} for CDFSL. Finally, we examine the distinct issues and challenges posed by CDFSL.

\vspace{-0.2cm}
\subsubsection {Empirical Risk Minimization (ERM)~\cite{erm1,erm2}}
Given an input space $\mathcal{X}$ and label space $\mathcal{Y}$, in which $X$ and $Y$ satisfy the joint probability distribution $P(X,Y)$, a loss function $l(\hat{y},y)$, a hypothesis $h \in \mathcal{H}$~\footnote{Hypothesis space $\mathcal{H}$ consists of all functions that can be represented by some choice of values for the weights~\cite{ml}. A hypothesis $h$ is a function in Hypothesis space.}, the risk (expected risk) of hypothesis $h(x)$ is defined as the expected value of the loss function:
\begin{align}
R(h)=\mathbb{E}[l(h(x),y)]=\int l(h(x),y)dP(x,y),
\end{align}
The ultimate goal of the learning algorithm is to find the hypothesis $h^{\ast}$ that minimizes the risk $R(h)$ in the hypothesis space $\mathcal{H}$:
\begin{align}
h^{\ast}=\text{argmin}_{h\in \mathcal{H}}R(h),
\end{align}
Since $P(x,y)$ is unknown, we compute an approximation of $R(h)$ called empirical risk by averaging the loss function over the training set:
\begin{align}
\hat R(h)=\frac{1}{n}\sum_{i=1}^{n}l(h(x_{i}),y_{i}),
\end{align}
Therefore, the expected risk is usually infinitely approximated by empirical risk minimization ~\cite{erm1,erm2}, \ie, a hypothesis $\hat{h}$ is chosen to minimize the empirical risk:
\begin{align}
\hat{h}=\text{argmin}_{h\in \mathcal{H}}\hat{R}(h).
\end{align}

In FSL, due to limited supervised data, the empirical risk $\hat R(h)$ may poorly approximate the expected risk $h^{\ast}$, leading to overfitting of the empirical risk minimization hypothesis $\hat{h}$. In other words, the core problem of FSL is the unreliable empirical risk caused by insufficient supervised data~\cite{fslsurvey}. Current FSL approaches typically address this overfitting by incorporating additional datasets and tasks. However, as tasks differ between source and target domains, FSL also faces knowledge transfer challenges due to task shift. This is illustrated in the following two-stage empirical risk minimization problem.

\vspace{-0.2cm}
\subsubsection{Two-Stage Empirical Risk Minimization (TSERM)}
We assume that all tasks are different but related, and we extend the two-stage empirical risk minimization (TSERM) from~\cite{tltheory} to explain the CDFSL problem. TSERM aims to transfer prior knowledge from the source task to the target task. In the first stage, the focus is on learning the prior knowledge. The second stage then uses this learned prior knowledge to construct an optimal hypothesis for the target task.

\begin{figure}
	\centering
  \vspace{-0.3cm}
	\includegraphics[width=\linewidth]{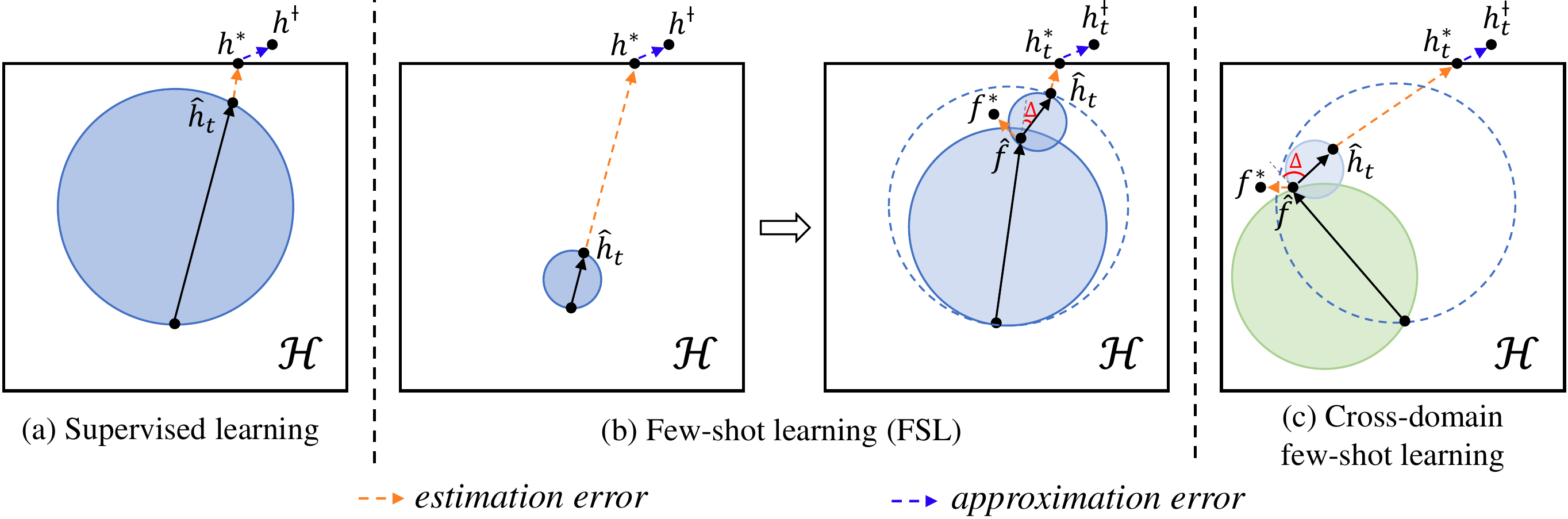}
 \vspace{-0.4cm}
	\caption{Comparison of (a) vanilla supervised learning, (b) few-shot learning (FSL), and (c) cross-domain few-shot learning (CDFSL). The square represents the hypothesis space $\mathcal{H}$. Solid circles denote the datasets (the size means the amount of data, \ie $\mathcal{D}$), the large and small solid circles represent the auxiliary and limited target datasets, respectively. Dotted circles indicate the domain to which the target samples belongs, which means the auxiliary dataset is from the same domain with target domain in (b), and different but related domain in (c). The angle between the optimization directions of the two stages represents the difference between the source and target tasks $\Delta$, \ie, the larger the angle, the greater the difference.}
 \vspace{-0.2cm}
	\label{issue1}
\end{figure}

Specifically, we use $\mathcal{D}^{s}$ and $\mathcal{D}^{t}$ to indicate the source and target domains, $\mathcal{T}^{s}$ and $\mathcal{T}^{t}$ to represent the source task and target task. TSERM learns two hypotheses $f$ and $h$~\footnote{both $f$ and $h$ are parametric models due to only limited supervised samples existing} in a hypothesis space $\mathcal{H}$, where $f$ learns the prior knowledge in the first stage, and $h$ utilizes it to adapt $\mathcal{T}^{t}$ in the second stage. For convenience, we use
\begin{itemize}
\item [(1)] $(h^{\dagger},f^{\ast})$~\footnote{we assume that there exists a common nonlinear feature representation $f^{\dagger}$ in $\mathcal{H}$, which means $f^{\ast}$=$f^{\dagger}$}=$\text{argmin}_{(f,h)}R(h, f)$ represents the function that minimizes the expected risk,

\item [(2)] $(h^{\ast},f^{\ast})$ =
$\text{argmin}_{(f,h)\in \mathcal{H}}R(h, f)$ means the function that minimizes the expected risk in $\mathcal{H}$,

\item [(3)] $(\hat{h},\hat{f})=
\text{argmin}_{(f,h)\in \mathcal{H}}\hat{R}(h, f)$ indicates the function that minimizes the empirical risk in $\mathcal{H}$.
\end{itemize}
Since $(h^{\dagger},f^{\ast})$ is unknown, it must be approximated by $(h, f)\in \mathcal{H}$. $(h^{\ast},f^{\ast})$ represents the most optimal approximation in $\mathcal{H}$, while $(\hat{h},\hat{f})$ represents the empirical risk minimization optimal hypothesis in $\mathcal{H}$. Suppose $(h^{\dagger},f^{\ast})$, $(h^{\ast},f^{\ast})$, $(\hat{h},\hat{f})$ are all unique. In the first stage, the empirical risk of $\mathcal{T}^{s}$ is given by the following formula:
\begin{align}
\hat{R}(h_s, f)=\frac{1}{N^{s}}\sum^{N^s}_{i=1}l(h_s\circ f(x_i^s),y_i^s),
\end{align}
where $l(\cdot, \cdot)$ is the loss function, $N^s$ represents the number of training samples in $\mathcal{D}^{s}$, and ($x_i^s$, $y_i^s$) represents the samples and corresponding labels in $\mathcal{D}^{s}$. $h_s$ is the hypothesis of $\mathcal{T}^{s}$. The prior knowledge extraction function $f^{'}(\cdot)$ is expressed as $(\hat{h}_{s},f^{'})=\text{argmin}_{(f,h_s)\in \mathcal{H}}\hat{R}(h_s,f)$. 
In the second stage, the empirical risk of $\mathcal{T}^{t}$ is defined as:
\begin{align}\hat{R}(h_t,f^{'})=\frac{1}{N^{t}}\sum^{N^t}_{i=1}l(h_t\circ f^{'}(x_i^t),y_i^t),
\end{align}
same as above, $h_t$ is the hypothesis of $\mathcal{T}^{t}$, $N^{t}$ denotes the number of training samples for $\mathcal{T}^{t}$, and $x_i^t$ and $y_i^t$ represent the samples and corresponding labels in $\mathcal{T}^{t}$, respectively. In the second stage, our goal is to estimate the optimal hypotheses $(\hat{h}_t,\hat{f})=\text{argmin}_{(f,h_t)\in \mathcal{H}}\hat{R}(h_t,f^{'})=\text{argmin}_{(f,h_t)\in \mathcal{H}}(\text{argmin}_{(f,h_s)\in \mathcal{H}}\hat{R}(h_s,f)+\lambda \Delta(\mathcal{T}^{s}, \mathcal{T}^{t}))$, where $\Delta(\mathcal{T}^{s}, \mathcal{T}^{t})$ means the distribution distance between $\mathcal{T}^{s}$ and $\mathcal{T}^{t}$, $\lambda$ represent the regularization parameter for weighting the distribution distance. We measure the function $(\hat{h_t},\hat{f})$ by the excess error~\cite{tltheory,fslsurvey} on $\mathcal{T}^{t}$, namely:
\begin{equation}
    \begin{aligned}
\mathbb{E}[R_{excess}] &  =\mathbb{E}[R(\hat{h_t},\hat{f})-R(h_t^{\dagger}, f^{\ast})] \\
& =\overbrace{\mathbb{E}[R(h^{\ast}_t,f^{\ast})-R(h^{\dagger}_ t,f^{\ast})]}^{\epsilon_{app}(\mathcal{H})}+\overbrace{\tilde{\mathbb{E}}[R(\hat{h}_t,\hat{f})-R(h^{\ast}_t,f^{\ast})]}^{\epsilon_{est}(\mathcal{H}, \mathcal{D}, \Delta)}.
\end{aligned}
\end{equation}
Among them, $R_{excess}$ represents the relationship between the expected risk of $(\hat{h_t},\hat{f})$ and the optimal prediction rule $(h_t^{\dagger},f^{\ast})$. The expectation $\mathbb{E}[\cdot]$ is with respect to the random choice of training data and training tasks. The approximation error $\epsilon_{app}(\mathcal{H})$ quantifies how well the functions in $\mathcal{H}$ approximate the optimal hypothesis $(h_t^{\dagger},f^{\ast})$. Meanwhile, the estimation error $\epsilon_{est}(\mathcal{H}, \mathcal{D}, \Delta)$ assesses the impact of minimizing the empirical risk $\hat{R}(h_t,f)$ in $\mathcal{H}$ instead of the expected risk $R(h_t,f)$, as shown by the orange dotted arrow in Figure \ref{issue1}. 

\vspace{-0.2cm}
\subsubsection{Unique Issue and Challenge} \label{challenge}
The approximation error $\epsilon_{app}(\mathcal{H})$, affected by $\mathcal{H}$, cannot be optimized due to the limitation of $\mathcal{H}$~\cite{fslsurvey}. Therefore, our goal is to optimize the estimation error $\epsilon_{est}(\mathcal{H}, \mathcal{D}, \Delta)$, which is affected by $\mathcal{H}$, amount of data in $\mathcal{D}$ (including $\mathcal{D}^{s}$ \& $\mathcal{D}^{t}$), and $\Delta$ ($\Delta$ is affected by the task shift and domain gap). This means $\epsilon_{est}(\mathcal{H}, \mathcal{D}, \Delta)$ can be reduced by increasing the amount of $\mathcal{D}$, constrain the complexity of $\mathcal{H}$, and having a small $\Delta$.

In Figure~\ref{issue1}, the solid black arrow expresses the learning of empirical risk minimization.  
Figure~\ref{issue1} (a) shows a vanilla supervised learning problem. It is easy to reduce $\epsilon_{est}(\mathcal{H}, \mathcal{D}, \Delta)$ in the case of large $\mathcal{D}$. The left part of (b) depicts the FSL problem, where ERM learning is suboptimal due to the insufficient data in $\mathcal{D}$. The existing FSL strategy, as shown in the right part of Figure \ref{issue1} (b), provides a large amount of auxiliary data from $\mathcal{D}$ and the corresponding different but relevant auxiliary task $\mathcal{T}^{s}$, here $\mathcal{T}^{s}\ne\mathcal{T}^{t}$, indicating only a task shift exists, \ie, $0 < \Delta \le \sigma$ where $\sigma$ is a small constant.

As a result of the domain gaps between the source and target datasets, a novel CDFSL problem emerges, as illustrated in Figure \ref{issue1} (c). It is evident that the CDFSL problem involves both domain gaps and task shift between the source and target domains, with limited supervised information available in $\mathcal{D}^{t}$. This means $\mathcal{D}^{t}$ is small and $\Delta$ is large in CDFSL, making $\epsilon_{est}(\mathcal{H}, \mathcal{D}, \Delta)$ large, leading to an unreliable TSERM (estimation error optimization) problem.

\vspace{-0.2cm}
\subsection{\textcolor{black}{Taxonomy}} \label{taxonomy}
\begin{figure}
	\centering
 \vspace{-0.2cm}
	\includegraphics[width=12cm]{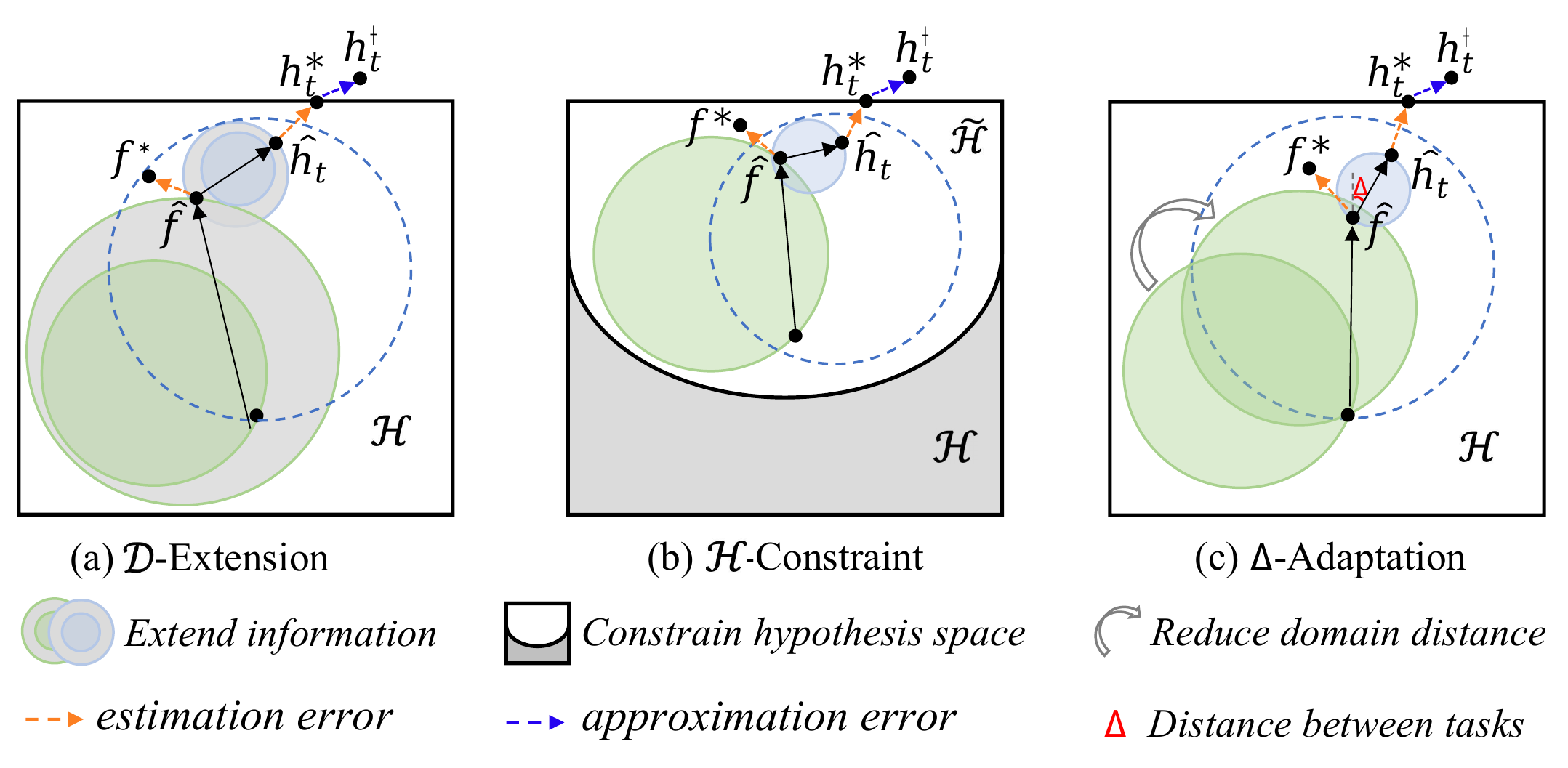}
  \vspace{-0.3cm}
	\caption{The main taxonomy of cross-domain few-shot learning (CDFSL) methods: (a) $\mathcal{D}$-Extension, (b) $\mathcal{H}$-Constraint, and (c) $\Delta$-Adaptation.}
 \vspace{-0.3cm}
	\label{step2}
\end{figure}

\begin{figure}[h]
	\centering
  \vspace{-0.3cm}
	\includegraphics[width=\linewidth]{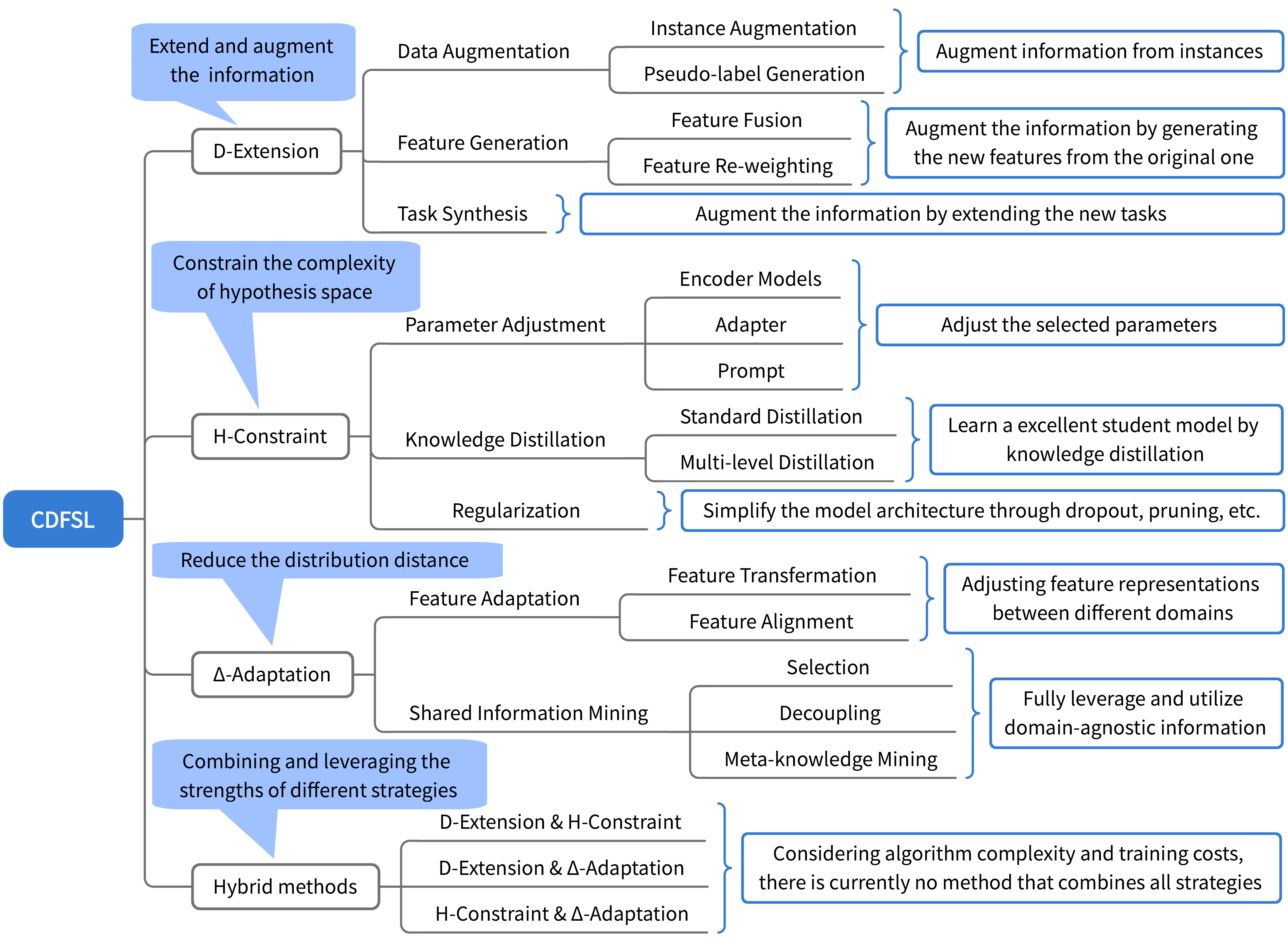}
 \vspace{-0.3cm}
	\caption{The taxonomy of representative methods in CDFSL.}
 \vspace{-0.4cm}
	\label{out}
\end{figure}

According to the unique issues and challenges mentioned above, CDFSL aims to solve the  unreliable TSERM problem. Therefore, existing CDFSL methods address this issue, as shown in Figure~\ref{step2}, through: (1) $\mathcal{D}$-extension, \ie augmenting the information in $\mathcal{D}$, (2) $\mathcal{H}$-constraint, \ie constraining the complexity of the hypothesis space $\mathcal{H}$, (3) $\Delta$-adaptation, \ie reducing the distance $\Delta$ between $\mathcal{T}^{s}$ and $\mathcal{T}^{t}$, and (4) Hybrid approaches, \ie combining the above strategies.
Accordingly, existing works are categorized into a unified taxonomy, as shown in Figure~\ref{out}. In the following sections, we will detail each category, performances, future works, and conclusion.

\vspace{-0.2cm}
\section{Approaches} \label{methods}
\begin{figure}
	\centering
	\includegraphics[width=11cm]{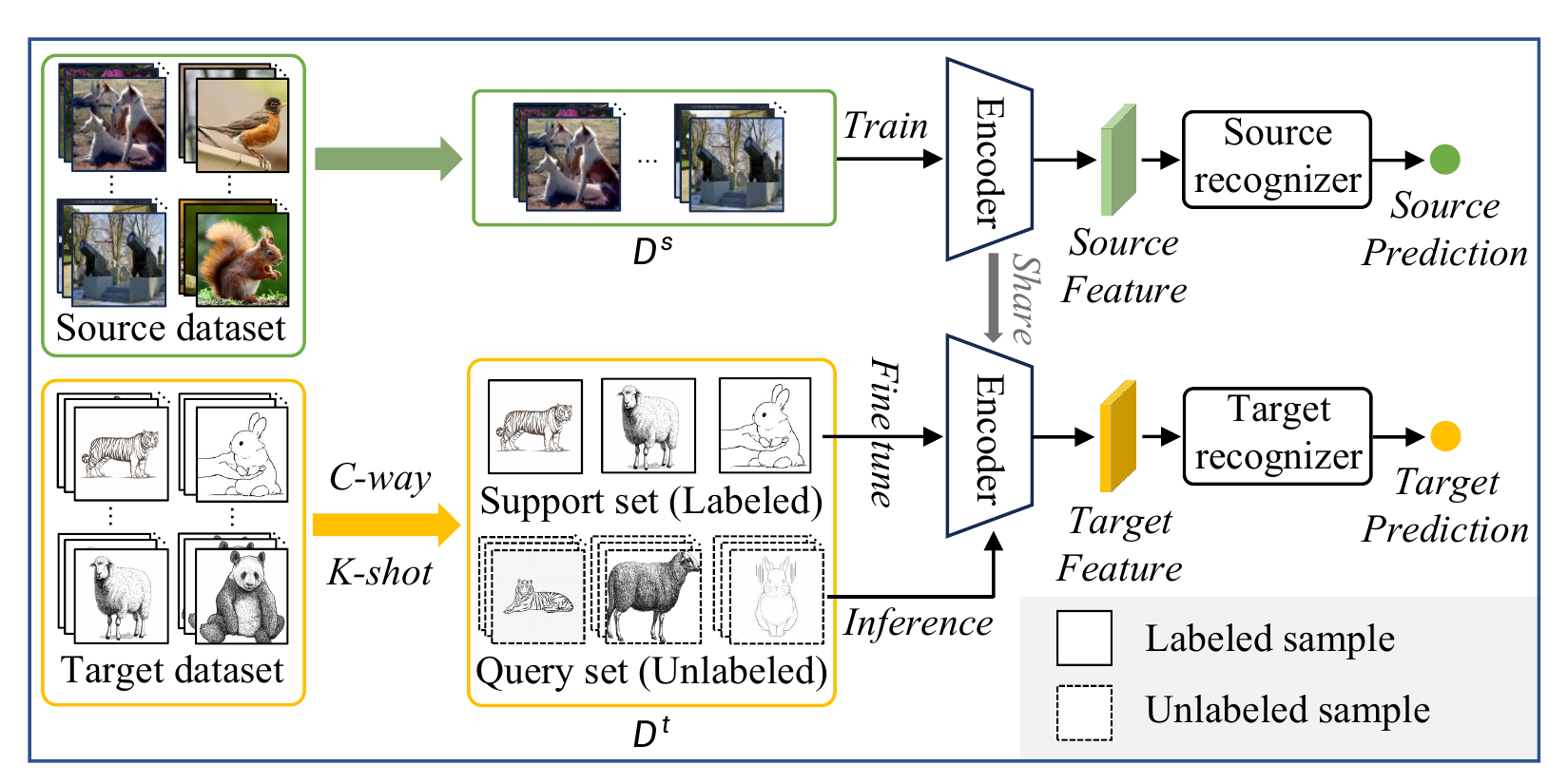}
  \vspace{-0.3cm}
	\caption{An overview of the CDFSL method involves two main stages. First, existing techniques pretrain the encoder on the source domain. Second, the encoder is finetuned, and a target recognizer is trained on the support set (with limited labeled data) of target domain. And inference is performed using the finetuned encoder and the target recognizer on the query set. $\textit{D}^{s}$ and $\textit{D}^{t}$ mean the source and target data.}
  \vspace{-0.3cm}
	\label{overview}
\end{figure}

CDFSL offers a unified solution to both cross-domain and FSL problems. Figure \ref{overview} outlines the CDFSL baseline process, consisting of two key steps: (1) Training a feature extractor on the source domain, and (2) Performing FSL on the target domain. The first step provides prior knowledge from the source domain, optimized via transfer learning~\cite{tfsurvey}, meta-learning~\cite{hospedales2021meta}, and metric learning~\cite{kaya2019deep} \etc techniques. The second step focuses on learning the feature extractor and target classifier with limited supervision. Based on the unique challenges, we categorize CDFSL algorithms into four types: $\mathcal{D}$-Extension, $\mathcal{H}$-Constraint, $\Delta$-Adaptation, and Hybrid Methods.

\vspace{-0.2cm}
\subsection{$\mathcal{D}$-Extension}  \label{instance}
\begin{figure}
	\centering
	\includegraphics[width=\linewidth]{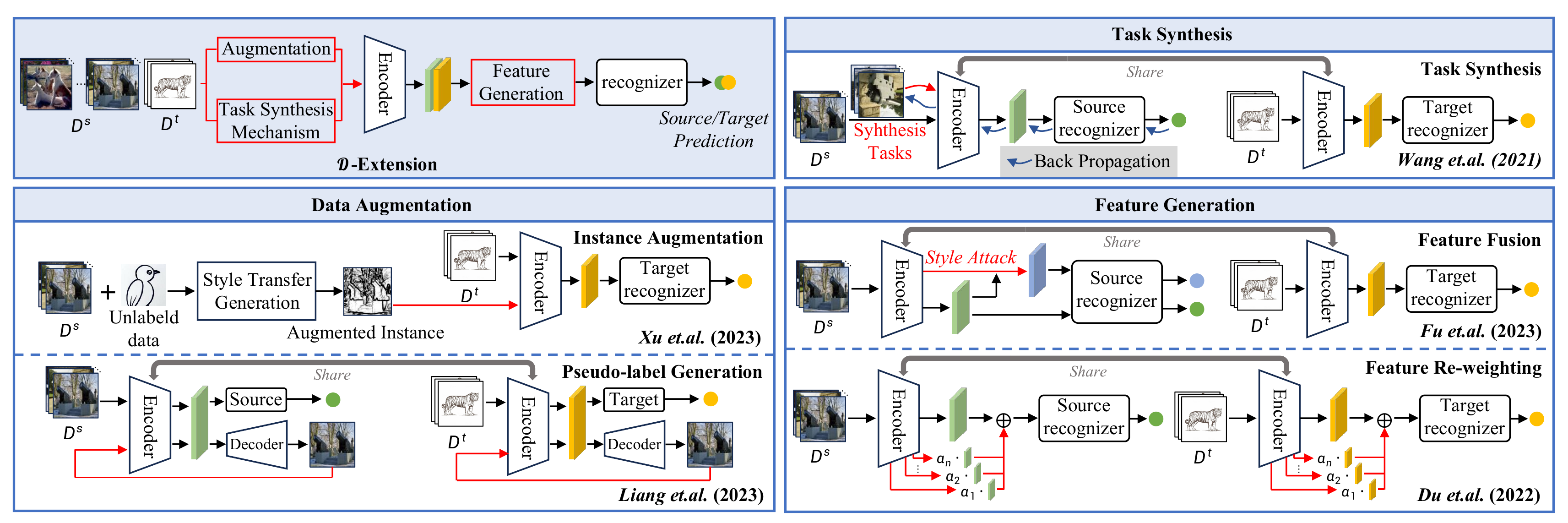}
  \vspace{-0.5cm}
	\caption{Solving CDFSL problem by $\mathcal{D}$-Extension, including data augmentation, feature generation, and task synthesis. Data augmentation aims to augment data by instance/pseudo-label generation, \etc. Feature generation enhances the features by fusing or re-weighting them. And task synthesis improves model generalization by generating more guided tasks. The core operations of methods are highlighted in red.
    }
  \vspace{-0.4cm}
	\label{fig_instance}
\end{figure}

$\mathcal{D}$-Extension enhances and expands existing information resources to improve the model’s generalization to new tasks, addressing the issue of unreliable TSERM. In CDFSL, information is typically represented in three dimensions: data space, feature space, and task space. Based on the dimension of information expansion, methods are classified into data augmentation, feature generation, and task synthesis. This classification demonstrates how $\mathcal{D}$-Extension systematically addresses the challenge of information scarcity across these three dimensions. Figure~\ref{fig_instance} illustrates how these methods work.

\vspace{-0.2cm}
\subsubsection{Data Augmentation.} \label{data_original}
Data augmentation expands the information by generating additional samples. Current methods introduce various types of transformations for models to learn, such as flipping~\cite{shyam2017attentive}, scaling~\cite{lake2015human, zhang2018fine}, translation~\cite{benaim2018one, lake2015human}, cropping~\cite{zhang2018fine}, shearing~\cite{shyam2017attentive}, reflection~\cite{edwards2017towards}, and rotation~\cite{model32, miniimagenet}. However, it's impractical to account for all potential transformations manually. Thus, affine transformations alone cannot fully solve the CDFSL problem~\cite{shyam2017attentive, lake2015human}. Next, we explore more advanced data augmentation methods at the sample and label levels, including instance augmentation and pseudo-label generation.

\textbf{Instance Augmentation.}
Instance augmentation~\cite{data_multi_3,sreenivas2023similar,extend_1,data_other_1,adcd,boosting,tgdm,liu2024spectral} increases sample diversity by synthesizing new data. Techniques include style transfer~\cite{data_multi_3,sreenivas2023similar}, frequency domain transformation~\cite{extend_1,data_other_1,liu2024spectral}, and adversarial methods~\cite{adcd,boosting}. For instance, \cite{data_multi_3} introduces unlabeled data from multiple domains into the original source domain, generating large amounts of data with different styles but the same content. \cite{sreenivas2023similar} augments data by applying the style of semantically similar categories to each class. Additionally, \cite{extend_1} enhances samples by processing high-frequency components of source images, allowing the network to mimic human visual perception by selecting different frequency cues in recognition tasks. Similarly, \cite{data_other_1} generates a sketch map from the original image to capture high-frequency information. \cite{liu2024spectral} explores a data augmentation method that augments the image in the frequency domain and then converts it back to the spatial domain for use as synthetic data. \cite{adcd} improves model robustness by generating adversarial instances, while \cite{tgdm} facilitates knowledge transfer by introducing an intermediate domain through mixing source and target domain images. Furthermore,~\cite{boosting} performs adversarial training with noise data, enhancing model robustness.

\textbf{Pseudo-label Generation.}
Pseudo-label generation involves leveraging unlabeled data by assigning pseudo labels to it~\cite{data_target_2,st}. One early approach~\cite{data_target_2} enhances unlabeled data by applying rotations and assigning the rotation degree as the pseudo labels, turning transformation tasks into supervised learning problems. Additionally,~\cite{st} introduces unlabeled target data during the pre-training stage, generating pseudo labels for these data through a pre-trained model. This unlabeled target data serves as supplementary information from the target domain, helping to enrich the model's understanding of the target domain without manual annotation. Following~\cite{st},~\cite{samarasinghe2023cdfsl} addresses the cross-domain few-shot video action recognition.

\vspace{-0.2cm}
\subsubsection{Feature Generation} \label{data_multiple}
Feature generation, as shown in Figure~\ref{fig_instance}, creates new feature representations from existing ones, enriching the feature space and improving the model's ability to capture data patterns, especially in scenarios with limited data. The goal of generating new features is either to expand the total information content or to explore key features in greater depth. According to the focus of generating features, methods can be divided into two categories: (1) feature fusion~\cite{rf,feature_fusion_2,mdfsl,parameter_fix_3,feature_reweight_6}, aims to increase the total information content, and (2) feature re-weighting~\cite{parameter_weight_2,ji2024relevance,parameter_fix_9,feature_reweight_3,feature_reweight_7,parameter_weight_3,adversarial,feature_reweight_0}, which focuses on increasing the amount of critical information in the feature to optimize information usage.

\textbf{Feature Fusion.}
There are two main types of feature fusion methods~\cite{rf,feature_fusion_2,mdfsl,parameter_fix_3,feature_reweight_6} discussed here: multi-level fusion~\cite{rf,feature_fusion_2} and multi-model fusion~\cite{mdfsl,parameter_fix_3,feature_reweight_6}. Multi-level fusion focuses on combining features from different layers of the same model, addressing the limitations of using single-layer features. Low-level features may lack the specificity needed for complex tasks, while high-level features can overfit to the training data. By fusing these features, a more balanced and versatile representation is created, capable of handling a wide range of tasks~\cite{lin2017feature}. For example,\cite{rf} computes the loss for the last few layers of the model and sums them up as the final loss, while\cite{feature_fusion_2} fuses mid-level features to predict the residual term.
Multi-model fusion~\cite{mdfsl,parameter_fix_3,feature_reweight_6} refers to combining features generated by different models or strategies. Since these features come from diverse sources, fusing them leverages the strengths of each, creating a richer and more distinguishable feature representation. For instance,\cite{mdfsl} fuses knowledge from both class and semantic aspects,\cite{parameter_fix_3} averages the features from multiple models, and~\cite{feature_reweight_6} constructs a non-linear subspace, then combines the features from this subspace with the original features.

\textbf{Feature Re-weighting.}
Feature re-weighting~\cite{parameter_weight_2,ji2024relevance,parameter_fix_9,feature_reweight_3,feature_reweight_7,parameter_weight_3,feature_reweight_0,adversarial} adjusts the weights of different features so the model can adaptively select the most important ones for different tasks or domains, improving performance. For example, \cite{parameter_weight_2} learns separate weights for each multi-level feature, \cite{parameter_fix_9} re-weights features across different episodes, and~\cite{feature_reweight_7} introduces a forget-update module to re-weight features. Meanwhile, \cite{parameter_weight_3} applies attention mechanisms across multiple levels, and \cite{ji2024relevance} re-weights both original and style-transformed features through context interaction. \cite{feature_reweight_3} enhances feature representation by leveraging local feature correlations, while \cite{feature_reweight_0} inputs style-altered features into the FSL classifier.~\cite{adversarial} explores adversarial feature augmentation to simulate distribution variations by re-weighting features.

\vspace{-0.2cm}
\subsubsection{Task Synthesis} \label{data_target}
Task synthesis~\cite{ata,feature_fusion_3,rao2024rdprotofusion,data_other_2} (Figure~\ref{fig_instance}) expands the task space by creating new task scenarios or combinations, improving the model’s ability to generalize. Some methods generate challenging tasks to train the model in difficult situations. For example, \cite{ata} generates tasks simulating worst-case scenarios, helping the model adapt to a wider distribution of tasks during training. Other approaches synthesize tasks at the graph level. 
For instance, \cite{feature_fusion_3} present few-shot data as graphs and augment them from contextual and geometric perspectives, while \cite{rao2024rdprotofusion} integrates information from different tasks using a multi-task learning framework. Additionally, \cite{data_other_2} randomly combines affine-transformed samples into diverse tasks.

\vspace{-0.2cm}
\subsubsection{Discussion and Summary}
When additional unsupervised samples or data generation methods are available, the amount of information can be expanded through data augmentation. However, in data augmentation, particularly in instance augmentation, the augmented information is typically derived from the original samples, which results in limited valid information gain and a lack of real-world diversity. Additionally, the quality of generated pseudo-labels depends on the model's predictions; if the model has biases, the pseudo-label generation process may further amplify these biases.

When multiple features, including those from multiple models or levels, are available, feature generation methods can be applied. These methods enrich the feature space by processing multiple features to enhance the information. However, since these features come from the same examples, they are often highly correlated or redundant, which increases model complexity while providing limited additional information. Therefore, finding the optimal solution requires very fine adjustment~\cite{he2016deep}.

Finally, when auxiliary tasks are present, task synthesis enhances diversity by combining new tasks. However, this increased diversity can make the model’s learning objectives unclear and task relevance ambiguous. This occurs because different tasks may conflict or lack consistency, making it difficult for the model to identify which features or patterns are most important for the primary task. Additionally, auxiliary tasks can introduce irrelevant noise, further confusing the learning process. As a result, the composed tasks may not align well with the actual task, leading to sub-optimal performance in real-world applications~\cite{al32}.

\vspace{-0.2cm}
\subsection{$\mathcal{H}$-Constraint}  \label{hypothesis}
\begin{figure}
	\centering
  \vspace{-0.3cm}
	\includegraphics[width=\linewidth]{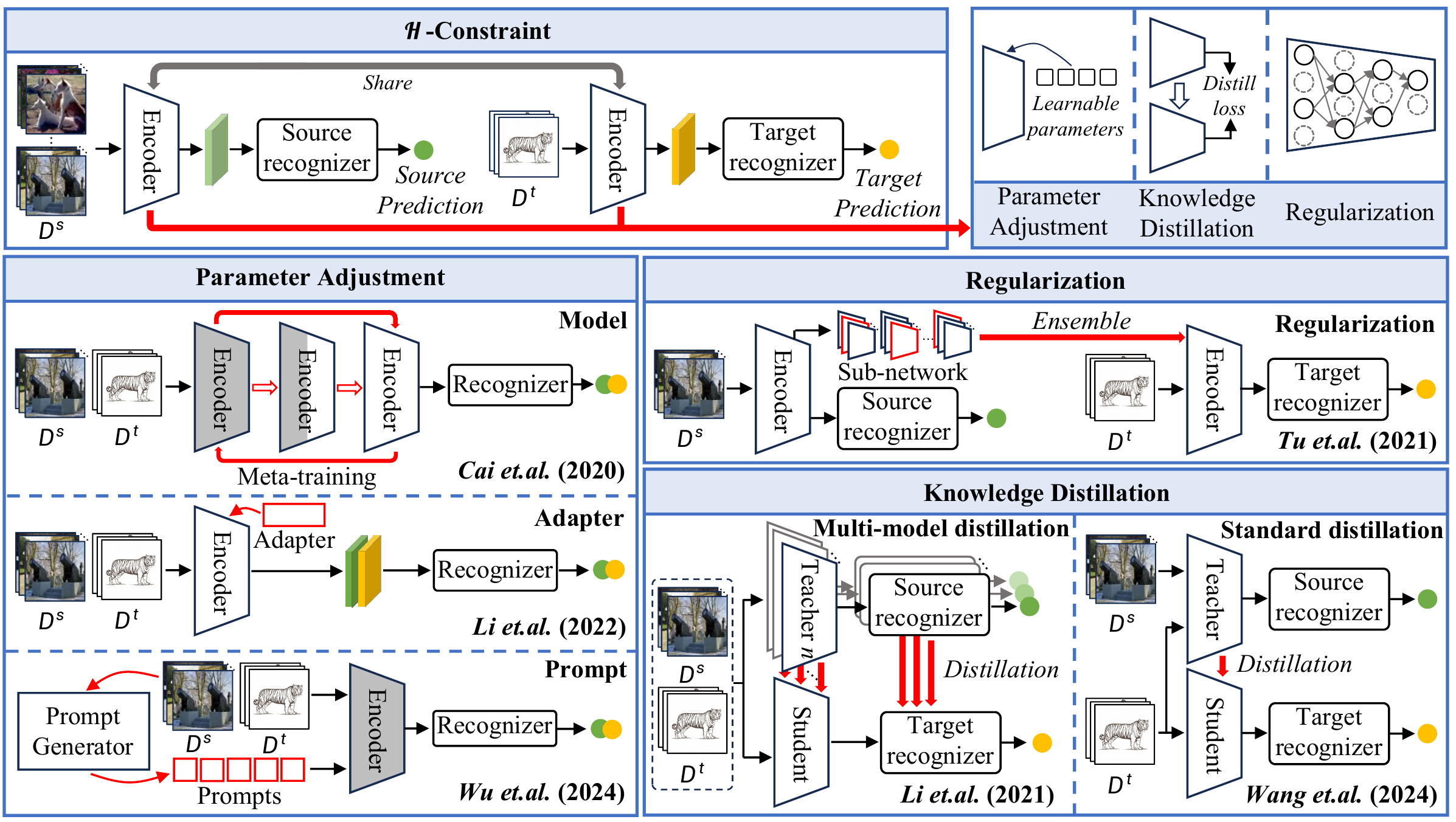}
 \vspace{-0.5cm}
	\caption{$\mathcal{H}$-Constraint methods, including parameter adjustment, knowledge distillation, and regularization. The gray background means the frozen parameters, and the core operations of methods are highlighted in red.}
 \vspace{-0.4cm}
	\label{parameterr}
\end{figure}

$\mathcal{H}$-Constraint are designed to solve the problem of unreliable TSERM by reducing the complexity of the hypothesis space $\mathcal{H}$. A smaller $\mathcal{H}$ limits the model’s complexity, which helps reduce overfitting and improves performance in the target domain. The methods in this section explore constraints on $\mathcal{H}$ from three perspectives: structural constraints, guided constraints, and normative constraints. Based on how these constraints are implemented, the methods are categorized into three types: (1) parameter adjustment, (2) knowledge distillation, and (3) regularization, as illustrated in Figure~\ref{parameterr}.

\vspace{-0.2cm}
\subsubsection{Parameter Adjustment} \label{fixed}
Parameter adjustment is a common strategy for imposing structural constraints, directly tuning parameters to reduce the complexity of the hypothesis space~\cite{parameter_fix_1,parameter_weight_1,parameter_weight_4,guo2023task,wu2024task,wu2024hybridprompt,gondal2024domain,zhuo2024prompt}. These parameters can come from existing models or be introduced through additional modules. Based on the source of the adjusted parameters, methods in this strategy are categorized into three groups: encoder models~\cite{parameter_fix_1,parameter_weight_1}, extra adapters~\cite{parameter_weight_4,guo2023task,gondal2024domain}, and prompts~\cite{wu2024task,wu2024hybridprompt,zhuo2024prompt,xu2024step}.

\textbf{Encoder Models.}
A common method for parameter adjustment is through the encoder models~\cite{parameter_fix_1,parameter_weight_1}. For instance,~\cite{parameter_fix_1} constrains the hypothesis space by alternately freezing certain model parameters in the inner and outer loops, allowing parameters in different regions to maximize their effectiveness. In addition,~\cite{parameter_weight_1} resets the parameters learned from the source domain before adapting to the target data. However, directly adjusting the parameters of the encoder model strictly limits the number of parameters that can be adjusted. When the encoder model has a large number of parameters, this strategy becomes inefficient.

\textbf{Adapters.}
To reduce the costs of parameter adjustment, extra adapters have been explored for insertion during the training phase~\cite{parameter_weight_4,guo2023task,gondal2024domain}. Specifically,~\cite{parameter_weight_4} constrains $\mathcal{H}$ by dynamically adapting the network via task-specific adapters. ~\cite{guo2023task} adjusts parameters through dynamic selection and activation of these task-specific adapters. And~\cite{gondal2024domain} introduces a lightweight adapter that is specifically trained
with an intra-modal contrastive objective.

\textbf{Prompts.}
With the widespread application of prompt learning, some methods~\cite{wu2024task,wu2024hybridprompt,zhuo2024prompt,xu2024step} address the CDFSL problem by adjusting only the prompt parameters. For example, \cite{wu2024task} generates task-adaptive prompts through a network, which are then used to guide the backbone network in solving the FSL task on the target domain. Building on this, \cite{wu2024hybridprompt} explores domain-aware prompting, transferring prompts learned from the source task to the target task. Moreover, \cite{zhuo2024prompt} treats prompts as a ``free lunch'' to enhance diversity and utilizes textual modality to guide prompt generation. And \cite{xu2024step} proposes and optimizes a style prompt to reduce the domain gaps.

\vspace{-0.2cm}
\subsubsection{Knowledge Distillation} \label{pa-select}
Knowledge distillation introduces additional supervisory signals through the soft labels, guiding the constraint on $\mathcal{H}$~\cite{wang2024cross,hybrid_4,yang2024leveraging,hybrid_5,data_target,data_multi_0}. Specifically, knowledge distillation involves using the outputs of a pre-trained teacher model as soft labels to guide the training of a student model, effectively transferring the teacher model's knowledge to the student model. We classify these methods from a technical perspective into standard knowledge distillation and multi-level knowledge distillation. For the standard knowledge distillation,~\cite{wang2024cross} uses a teacher-student network, combined with self-supervised learning, to generate soft labels that guide the training of the student model.

Other methods~\cite{hybrid_4,data_multi_0,yang2024leveraging,hybrid_5,data_target} explore multi-level knowledge distillation to solve the CDFSL problem. Specifically, there are multi-to-one and one-to-multi teacher-student networks. For example,~\cite{hybrid_4} trains a student network with multiple teacher networks, where these teacher networks are trained on datasets from different domains, respectively. ~\cite{hybrid_5} aligns the representations of the single student network with the multiple task/domain-specific ones using low-capacity adapters. Additionally,~\cite{data_target} guides the learning of the student model through two teacher models, including the source and the target teacher models. ~\cite{data_multi_0} proposes a multi-level knowledge stage to train the student network using multiple teacher models. In contrast,~\cite{yang2024leveraging} trains multiple student adapters in different stages, with the help of a single teacher network.

\vspace{-0.2cm}
\subsubsection{Regularization} \label{pa-weight}
Regularization constrains $\mathcal{H}$ by adding penalty terms to the loss function or introducing randomness, such as dropout. Some CDFSL approaches~\cite{feature_reweight_1,ji2023cross,parameter_select_1} apply regularization to limit $\mathcal{H}$. For example, \cite{feature_reweight_1} uses explainable feedback to guide model training, while \cite{ji2023cross,parameter_select_1} introduce dropout during training to impose constraints. Specifically, \cite{parameter_select_1} samples sub-networks by dropping individual neurons or entire feature maps, then selects the most suitable sub-networks to form an ensemble for target domain learning. Additionally, \cite{ji2023cross} combines pruning with dropout to dynamically adjust the model's complexity.

\vspace{-0.2cm}
\subsubsection{Discussion and Summary}
Alternating the adjustment of encoder model parameters allows for optimization across different tasks and datasets, improving model performance. However, this requires a well-designed strategy to avoid local optima, as an inappropriate adjustment strategy may lead to overfitting and reduced generalization~\cite{srivastava2014dropout}. Extra adapters and prompts can be flexibly inserted into existing models~\cite{houlsby2019parameter} to facilitate parameter updates, and prompt learning can achieve better performance with limited training data~\cite{gao2021making}. However, the design and selection of prompts are critical, as each task requires tailored prompts for optimal performance~\cite{brown2020language}.

For the knowledge distillation strategy (Section~\ref{pa-select}), by learning from the soft labels generated by the teacher model, the student model can better handle noisy data and label errors, thereby improving its robustness. However, the effectiveness of knowledge distillation depends on the quality of the teacher model—if the teacher model performs poorly, the student model will also be negatively affected~\cite{hinton2015distilling}. Additionally, knowledge distillation requires extra training steps, increasing both the complexity and duration of the training process~\cite{gou2021knowledge}.

Regularization (Section~\ref{pa-weight}) constrains the magnitude of weights, helping the model generalize better to unseen data~\cite{goodfellow2016deep}. However, in scenarios like small sample learning, regularization may not significantly improve performance~\cite{ng2004feature}. Additionally, if regularization parameters are not properly selected, such as when the regularization strength is too high—the model may be over-constrained, limiting its ability to learn complex patterns. In such cases, overfitting may not be effectively prevented.

\vspace{-0.2cm}
\subsection{$\Delta$-Adaptation}  \label{adaptation}
\begin{figure}
	\centering
  \vspace{-0.3cm}
 \includegraphics[width=\linewidth]{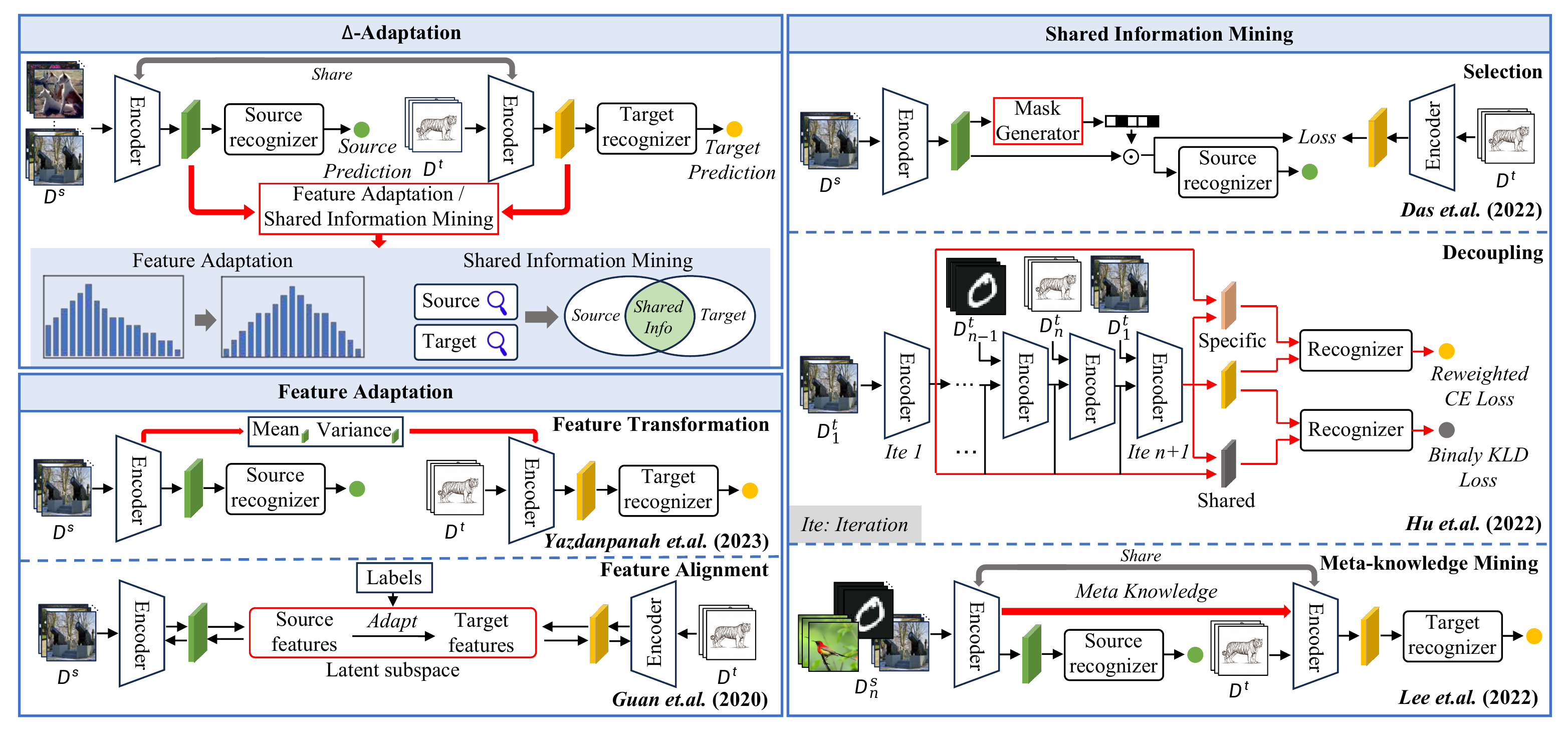}
 \vspace{-0.6cm}
	\caption{The $\Delta$-Adaptation categories, including feature adaptation and shared information mining. Feature adaptation involves feature transformation and feature alignment. Shared information mining involves selection, decoupling, and meta-knowledge mining. The core operations of methods are highlighted in red.}
 \vspace{-0.4cm}
	\label{feature-post}
\end{figure}

In FSL, source and target tasks usually come from the same domain, meaning $\Delta$ is nearly zero. However, in CDFSL, the gap between the source and target tasks can be much larger, leading to an unreliable TSERM. To address this, it’s important to reduce the disparity $\Delta$ between the tasks. This can be done in two ways: (1) adapting the features of both tasks to minimize differences, or (2) exploring and leveraging shared information to improve robustness even when differences remain. Based on how $\Delta$ is optimized, methods in this section are categorized into two groups (Figure~\ref{feature-post}): (1) Feature adaptation and (2) Shared information mining.

\vspace{-0.2cm}
\subsubsection{Feature Adaptation} \label{select}
Feature adaptation aims to reduce the differences between the source and target domains by adjusting or transforming their features. Since the feature forms or structures of the source and target domains may exhibit fundamental differences, these disparities are often difficult to reconcile directly in the original feature space. Therefore, depending on whether the original feature space is preserved, the methods in this section are categorized into two approaches: feature transformation~\cite{feature_reweight_4,parameter_fix_8,wang2023mmt,feature_reweight_2,liu2022geometric,li2023knowledge}, which modifies the feature space, and feature alignment~\cite{paeedeh2024cross,feature_reweight_9,lscdfsl,zhao2023dual,xu2024enhancing,feature_reweight_5,parameter_fix_6}, which attempts to align features without altering the original space.

\textbf{Feature Transformation.}
Feature transformation~\cite{feature_reweight_4,parameter_fix_8,wang2023mmt,feature_reweight_2,liu2022geometric,li2023knowledge} aligns the features of the source and target domains by applying transformations to the original feature space. For instance, \cite{feature_reweight_4} enhances CDFSL performance by using adaptive feature distribution transformations and task-adaptive strategies to measure domain discrepancies. \cite{feature_reweight_2} generates inductive meta points to abstract and transform features, while~\cite{parameter_fix_8} applies scale and shift operations via a feature transformation module. \cite{wang2023mmt} uses memory components for effective feature transformation and alignment between domains. Similarly, \cite{liu2022geometric} maps features into a high-dimensional geometric algebra space to preserve relationships across domains, and~\cite{li2023knowledge} refines the model's understanding of domain-specific features, aligning learned features with the target domain's characteristics.

\textbf{Feature Alignment.}
Feature alignment~\cite{paeedeh2024cross,feature_reweight_9,lscdfsl,zhao2023dual,xu2024enhancing,feature_reweight_5,parameter_fix_6} aims to adjust the feature distributions of the source and target domains to achieve consistent representations. In~\cite{lscdfsl}, the features are mapped into a shared latent subspace to align source and target features. Similarly, \cite{paeedeh2024cross} achieves alignment using transformer blocks, while \cite{feature_reweight_9} learns cross-domain aligned representations through bi-directional feature alignment. \cite{xu2024enhancing} optimizes target domain features via information maximization, implicitly aligning features. Additionally, \cite{feature_reweight_5} uses instance normalization to reduce feature dissimilarity, and \cite{parameter_fix_6} finetunes Batch Normalization (BN) layers to correct inter-domain distribution differences. Finally,~\cite{zhao2023dual} combines prototypical feature alignment with normalized distribution alignment.

\vspace{-0.2cm}
\subsubsection{Shared Information Mining} \label{fuse}
Shared information mining focuses on efficiently leveraging common features, patterns, or knowledge between the source and target domains to enhance model generalization in cross-domain learning. These shared features must be both transferable across domains and relevant to the source and target tasks. Based on the number of tasks involved in extracting shared information, existing methods are generally divided into three categories: selection, decoupling, and meta-knowledge mining.

\textbf{Selection.}
Selection operates within a single task, focusing on filtering features to identify the most relevant information~\cite{feature_select_1,feature_select_2,confess,data_multi_1,parameter_select_2,zhang2023cross}. The selected information is then used as shared knowledge. Methods in this section are classified into feature selection, parameter weighting, and task optimization. Feature selection~\cite{feature_select_1,feature_select_2,confess} identifies general feature components that aid the target FSL task. For instance, \cite{feature_select_1} proposes a multi-domain feature selection algorithm to optimize feature extraction and selection, while \cite{feature_select_2} uses multiple feature extractors to select representations relevant to the target domain. Additionally, \cite{confess} uses a learnable mask generator to identify shared features. Parameter weighting \cite{data_multi_1,parameter_select_2} adjusts model parameters to adapt to the target task. For example, \cite{data_multi_1} aggregates meta-model parameters from multiple domains to generate initialization parameters, and \cite{parameter_select_2} uses a dynamic mechanism to configure the best adaptation modules for the downstream task. Finally, task optimization~\cite{zhang2023cross} focuses on selecting the best source tasks to learn valuable knowledge for the target task.

\textbf{Decoupling.}
In CDFSL, source and target task features share both similarities and differences. Decoupling~\cite{xu2023cross,data_target_1,feature_reweight_8,zhou2024meta} aims to separate shared information from task-specific information to handle them independently and avoid interference. In~\cite{xu2023cross}, class-shared and class-specific dictionaries are combined to align source and target features, with the class-shared dictionary capturing common elements and class-specific dictionaries representing domain-specific features. Similarly, \cite{data_target_1} employs domain-specific and domain-general encoders to support the learner, while \cite{feature_reweight_8} uses a domain discriminator to distinguish between source and target features, allowing the extractor to learn domain-invariant features and reduce domain discrepancies. In addition, \cite{zhou2024meta} decouples the image into low-frequency content details and high-frequency robust structural characteristics, and improves the prediction performance by combining the effective parts of two.

\textbf{Meta-knowledge Mining.}
Meta-knowledge mining~\cite{parameter_fix_5,data_multi_2} operates across tasks and domains, extracting higher-order learning strategies that enable models to quickly adapt to new contexts. In~\cite{parameter_fix_5}, meta-learning is applied across domain tasks during training to generate prototypes suited for new tasks. Similarly,~\cite{data_multi_2} introduces a domain-agnostic meta-learning method that learns domain-independent initial parameters by observing both seen and pseudo-unseen tasks simultaneously.

\vspace{-0.2cm}
\subsubsection{Discussion and Summary}
Feature adaptation strategies (Section~\ref{select}) typically achieve alignment by minimizing distribution differences between the source and target domains. These methods are flexible, easy to implement, and can be applied to various tasks and datasets. However, in focusing on matching overall distribution statistics (e.g., mean and variance), they often overemphasize global feature consistency, potentially overlooking the local or specific information crucial for accurately modeling the target domain~\cite{dasurvey1}.

Shared information mining seeks to identify and leverage common knowledge between domains (Section~\ref{fuse}) and can be applied to single-task, dual-task, and multi-task settings. However, its effectiveness may be compromised by conflicting task objectives~\cite{zhao2019learning}. For instance, features beneficial to one task may be irrelevant or harmful to another, potentially confusing the model during training. To mitigate this, some methods incorporate task-specific knowledge alongside shared information. Rather than directly applying shared features,\cite{data_target_1} constrains their use through domain-specific insights, while\cite{zhang2023cross} evaluates the impact of shared features on the target domain and leverages them accordingly.

\vspace{-0.3cm}
\subsection{Hybrid Methods}  \label{hybrid}
The three strategies above address the unreliability of TSERM by optimizing specific factors: $\mathcal{D}$-Extension enhances the available information in $\mathcal{D}$, $\mathcal{H}$-Constraint limits the complexity of the hypothesis space $\mathcal{H}$, and $\Delta$-Adaptation optimizes $\Delta$. Each strategy has distinct advantages: $\mathcal{D}$-Extension is simple and practical, $\mathcal{H}$-Constraint reduces model complexity, and $\Delta$-Adaptation improves the transferability of knowledge from the source to the target domain. As a result, many studies have explored combining these strategies to leverage their strengths and further mitigate TSERM unreliability. Related technologies are listed in Table~\ref{table_hybrid}.

Some methods combine $\mathcal{D}$-Extension and $\mathcal{H}$-Constraint techniques~\cite{dynamic,perera2024discriminative,cdfsl231,ji2024soft,hybrid_3}. For example, \cite{dynamic} introduces additional unsupervised target domain samples during the pre-training phase and augments them by training the target domain encoder through self-distillation. Similarly, \cite{cdfsl231} incorporates unlabeled target data during training, using cross-level knowledge distillation to enhance the model's capacity to extract highly discriminative features from the target dataset. \cite{perera2024discriminative} fuses features at different levels during pre-training and introduces learnable anchors (acting as adapters) during fine-tuning to adapt to the target task. Furthermore, \cite{hybrid_3} addresses the FSL task in the target domain by synthesizing tasks from multiple perspectives and applying regularization based on consistency and correlation metrics. Finally, following \cite{ji2023cross}, \cite{ji2024soft} introduces additional unlabeled target data and proposes an iterative soft pruning training method that combines contrastive loss and regularization to remove redundant weights and optimize the model.

Some methods combine $\mathcal{D}$-Extension and $\Delta$-Adaptation strategies~\cite{hybrid_1,hybrid_7,hybrid_2,feature_reweight_11,feature_reweight_10}. For instance, \cite{hybrid_1} introduces additional labeled target instances and decouples features into domain-agnostic and domain-specific categories during fine-tuning, leveraging domain-agnostic features with a domain and FSL classifier. As an extension of \cite{hybrid_1}, \cite{hybrid_7} uses a mixed module to generate diverse images from both source and target domains and decouples features during meta-training. \cite{hybrid_2} decouples features to obtain domain-agnostic and domain-specific components, then adaptively weights these features for the target task. \cite{feature_reweight_11} fuses features at different levels to enhance generalization, then adapts to the target task through feature transformation during meta-training. \cite{feature_reweight_10} synthesizes tasks using style transformations and employs these tasks to learn and optimize feature transformation modules.

In the combination of $\mathcal{H}$-Constraint and $\Delta$-Adaptation methods~\cite{feature-wise,parameter_fix_7,zhang2024cross,ma2023prod}, \cite{feature-wise} asynchronously freezes and updates the feature-wise transformation layers and feature extractor. Building on this, \cite{parameter_fix_7} introduces a metric function to better measure the distance between labeled and unlabeled images. Meanwhile, \cite{zhang2024cross} uses multi-step self-distillation to update a feature adaptation module at each learning step. In \cite{ma2023prod}, domain-agnostic and domain-specific prompts are designed to adapt to the target task.

\begin{table}
\tiny
\centering
\vspace{-0.2cm}
\caption{Representative hybrid CDFSL approaches.}
\vspace{-0.2cm}
\setlength{\tabcolsep}{0.6mm}{
\begin{tabular}{llccccccc}
\hline
\textbf{Methods} & \textbf{Venue} & \textbf{$\mathcal{D}$-Extension} & \textbf{$\mathcal{H}$-Constraint} & \textbf{$\Delta$-Adaptation} & \textbf{Loss function} & \textbf{FG} & \textbf{Art} & \textbf{IW}      \\ 
 \hline
DDN~\cite{dynamic}  & NIPS 2021 & Data Augmentation & Knowledge Distillation & \XSolidBrush & CE \& entropy &  &  & \Checkmark    \\
DIPA~\cite{perera2024discriminative}   &  CVPR 2024  & Feature Generation & Parameter Adjustment & \XSolidBrush & CE &  & \Checkmark &      \\
CLDFD~\cite{cdfsl231} & ICLR 2023 &  Data Augmentation & Knowledge Distillation & \XSolidBrush & CE \& SimCLR \& KD &  &  & \Checkmark    \\
SWP-LS~\cite{ji2024soft} & TMM 2024 & Data Augmentation & Regularization & \XSolidBrush & Contrastive \& $L_{2}$ & \Checkmark &  & \Checkmark \\  
TL-SS~\cite{hybrid_3}  &  AAAI 2022  & Data Augmentation & Parameter Adjustment & \XSolidBrush & CE \& Metric & \Checkmark &  & \Checkmark     \\  
Meta-FDMixup~\cite{hybrid_1} & ACM MM 2021 & Data Augmentation & \XSolidBrush & Shared Information Mining & CE \& KL & \Checkmark &   &     \\  
GMeta-FDMixup~\cite{hybrid_7} & TIP 2022 & Data Augmentation & \XSolidBrush & Shared Information Mining & CE \& SSL & \Checkmark &  & \Checkmark    \\  
Tri-M~\cite{hybrid_2}    &  ICCV 2021  & Feature Generation & \XSolidBrush & Shared Information Mining & CE & \Checkmark & \Checkmark &      \\
TCT-GCN~\cite{feature_reweight_11}    &  Neurocomputing 2023  & Feature Generation & \XSolidBrush & Feature Adaptation & CE & \Checkmark &  & \Checkmark     \\
TAML~\cite{feature_reweight_10} & arXiv 2023 & Task Synthesis & \XSolidBrush & Feature Adaptation & CE & \Checkmark &  & \Checkmark     \\
FWT~\cite{feature-wise} & ICLR 2020 & \XSolidBrush & Parameter Adjustment & Feature Adaptation & CE & \Checkmark &  &    \\
FGNN~\cite{parameter_fix_7} & KBS 2022 & \XSolidBrush & Parameter Adjustment & Feature Adaptation & Softmax & \Checkmark &  &    \\
FAD~\cite{zhang2024cross}    &  Neural Comput Appl 2024  & \XSolidBrush & Knowledge Distillation & Feature Adaptation & CE \& KL &  & \Checkmark &      \\
ProD~\cite{ma2023prod}    &  CVPR 2023  & \XSolidBrush & Parameter Adjustment & Shared Information Mining & CE \& Cos & \Checkmark &  &      \\
\hline
\end{tabular}}
\vspace{-0.3cm}
\label{table_hybrid}
\end{table}

\vspace{-0.2cm}
\subsubsection{Discussion and Summary}
The combination of multiple strategies in CDFSL, as discussed in Section~\ref{hybrid}, can lead to improved performance. For instance, due to its simplicity and ease of implementation, the $\mathcal{D}$-Extension strategy is often combined with other approaches. However, challenges arise when combining strategies. For example, combining $\mathcal{D}$-Extension and $\mathcal{H}$-Constraint methods can result in training conflicts, leading to negative transfer~\cite{zhao2024all}. Although $\mathcal{D}$-Extension increases data diversity, this diversity may not align with the target domain's characteristics, particularly when there is a significant feature distribution difference. In such cases, the features learned may be ineffective for the target domain. Conversely, $\mathcal{H}$-Constraint methods stabilize model performance by optimizing parameters, but during self-distillation, the model may become overly reliant on existing feature distributions. If these features don't match the target domain, negative transfer may occur~\cite{zhao2024all}.

Moreover, $\mathcal{D}$-Extension may generate additional noise, which negatively impacts methods combined with $\Delta$-Adaptation. This introduces irrelevant information during the process of reducing $\Delta$, potentially misleading the alignment or hindering the extraction of shared information, leading to performance degradation~\cite{liu2022deep}. To address this, some methods~\cite{hybrid_1,feature_reweight_0} control the augmentation process to reduce the introduction of irrelevant information.

The combination of $\mathcal{H}$-Constraint and $\Delta$-Adaptation has high training complexity and computational cost. Specifically, these two strategies require the model to simultaneously optimize the hypothesis space while aligning features between the source and target domains. $\mathcal{H}$-Constraint often involves intricate regularization techniques or model architecture modifications, while $\Delta$-Adaptation requires measuring and aligning distributional differences (\eg, through MMD or adversarial training). This combination increases the complexity of the optimization process, making training more demanding and significantly raising computational costs~\cite{liu2022deep,wang2020rethink}.
Currently, the high training complexity and computational cost prevent any method from addressing the TSERM unreliability problem by combining all three strategies. However, with the rise of large-scale pre-trained models (LMs) \cite{dosovitskiy2020image,radford2021learning,woo2023convnext} and prompt learning \cite{liu2023pre}, we expect future methods to explore this combination. The pre-training of LMs captures broad feature representations, reducing the need for extensive fine-tuning. At the same time, prompt learning provides a lightweight approach that leverages the knowledge in large models, minimizing the need for complex optimization and training costs.
\vspace{-0.2cm}
\section{Performance} \label{performance}
This section provides a comprehensive overview of the evaluation process for models in Cross-Domain Few-Shot Learning (CDFSL). To assess the effectiveness of these models, we first examine the relevant datasets and benchmarks in Section~\ref{benchmark}. In Section~\ref{per-com}, we offer a detailed analysis and comparison of the performance of various methods in CDFSL, highlighting the strengths and weaknesses of different approaches in tackling this complex problem.

\vspace{-0.2cm}
\subsection{Datasets \& Benchmarks} \label{benchmark}
The availability of annotated datasets enables fair comparison of CDFSL models and fosters the evaluation of various algorithms and architectures. The complexity, size, number of annotations, and transfer difficulty of these datasets present ongoing challenges that drive the development of innovative techniques. Table~\ref{dataset} lists the most widely used datasets for the CDFSL problem.
\begin{table}
\tiny
\centering
\vspace{-0.3cm}
\caption{Details of datasets in CDFSL.}
\vspace{-0.3cm}
\setlength{\tabcolsep}{1.2mm}{
\begin{tabular}{lcccccccc}
\hline
\textbf{Datasets} & \textbf{Derived from} & \textbf{Number of images} & \textbf{Image size} & \textbf{\makecell[c]{Number of \\ categories}} & \textbf{Content} & \textbf{Fields} & \textbf{Reference}     \\ 
 \hline
\textit{mini}ImageNet & ImageNet & 60000 & $84 \times 84$ & 100 & objects classification & natural scene & \cite{miniimagenet}    \\  
\textit{tiered}ImageNet & ImageNet & 779165 & $84 \times 84$ & 608 & objects classification & natural scene & \cite{tieredimagenet}     \\
Plantae  & iNat2017 & 196613 & varying & 2101 & plants \& animals classification & natural scene & \cite{plantae}  \\
Places     & N/A & 10 million & $200 \times 200$ & 400+ & scene classification  & natural scene & \cite{places}   \\
Stanford Cars     & N/A & 16185 & varying & 196 & cars fine-grained classification & natural scene &  \cite{cars}   \\
CUB   & ImageNet & 11788 & $84 \times 84$ & 200 & birds fine-grained classification & natural scene &  \cite{cub}   \\  
CropDiseases   & N/A & 87000 & $256 \times 256$ & 38 & crop leaves classification & natural scene &  \cite{cropdiseases}  \\
EuroSAT   & Sentinel-2 satellite & 27000 & $64 \times 64$ & 10 & land classification & remote sensing &  \cite{eurosat}   \\
ISIC 2018   & N/A & 11720 & $600 \times 450$ & 7 & dermoscopic lesion classification & medical &  \cite{isic1}   \\
ChestX   & N/A & 100K & $1024 \times 1024$ & 15 & lung diseases classification & medical & \cite{chestx}  \\
Omniglot   & N/A & 25260 & $28 \times 28$ & 1623 & characters classification & character &  \cite{omniglot}   \\
FGVC-Aircraft   & N/A & 10200 & varying & 100 & Aircraft fine-grained classification & natural scene &  \cite{aircraft}   \\
DTD   & N/A & 5640 & varying & 47 & textures classification & natural scene &  \cite{dtd}   \\
Quick Draw   & Quick draw! & 50 million & $128 \times 128$ & 345 & drawing images classification & Art &  \cite{draw1}   \\
Fungi   & N/A & 100000 & varying & 1394 & fungi fine-grained classification & natural scene &  \cite{fungi}   \\
VGG Flower   & N/A & 8189 & varying & 102 & flowers fine-grained classification & natural scene &  \cite{vgg}   \\
Traffic Signs   & N/A & 50000 & varying & 43 & Traffic signs classification & natural scene &  \cite{traffic}   \\
MSCOCO   & N/A & 1.5 million & varying & 80 & objects classification & natural scene &  \cite{mscoco}   \\
\hline
\end{tabular}}
\vspace{-0.5cm}
\label{dataset}
\end{table}

Based on these datasets, several benchmarks have been established for the CDFSL problem, including \textit{mini}ImageNet \& CUB (mini-CUB), the standard fine-grained classification benchmark (FGCB), and BSCD-FSL~\cite{bscd-fsl}. Additionally, Meta-Dataset~\cite{meta-dataset} has been proposed to evaluate the cross-domain problem in FSL. Since mini-CUB is included in FGCB, we mainly focus on introducing the last three benchmarks.

\textit{FGCB}.
An early benchmark for Fine-grained CDFSL ($FG$) was established during the initial development of CDFSL. It includes five datasets: \textit{mini}ImageNet, Plantae~\cite{plantae}, Places~\cite{places}, Cars~\cite{cars}, and CUB~\cite{cub}, with \textit{mini}ImageNet typically serving as the source domain and the others as the target domains. The benchmark consists entirely of natural images. The main challenge across domains for this benchmark is transferring the category information from coarse to fine.

\textit{BSCD-FSL}~\cite{bscd-fsl}.
As a more challenging benchmark for imaging way-based CDFSL ($IW$), BSCD-FSL includes five datasets: \textit{mini}ImageNet, CropDisease~\cite{cropdiseases}, EuroSAT~\cite{eurosat}, ISIC~\cite{isic1,isic2}, and ChestX~\cite{chestx}. Considering \textit{mini}ImageNet as the source domain, the distribution distances between the source and target datasets increase in the order: CropDiseases, EuroSAT, ISIC, and ChestX. BSCD-FSL is designed to evaluate methods across both near-domain and distant-domain scenarios.

\textit{Meta-Dataset}~\cite{meta-dataset}.
Meta-Dataset is a large-scale, diverse benchmark. It includes 10 publicly available datasets spanning natural images, handwritten characters, and graffiti, making it suitable for addressing Fine-grained ($FG$) and art-based ($Art$) CDFSL problems. This benchmark goes beyond the limitations of traditional \textit{C-way K-shot} tasks. It also introduces real-world class imbalances by varying the number of classes and training set sizes for each task.

Some methods use benchmarks originally designed for domain adaptation (DA). The DomainNet benchmark~\cite{domainnet}, addressing art-based cross-domain problems, includes 6 domains with 345 object categories. The Office-Home benchmark~\cite{officehome}, used in some CDFSL studies, comprises 4 domains (art, clipart, product, and real world) with 65 categories each and a total of 15,500 images, averaging 70 images per class.

\vspace{-0.2cm}
\subsection{Performance Comparison and Analysis} \label{per-com}
This section compares the performance of CDFSL approaches across various categories, using prediction accuracy as the standard metric under settings like \textit{5-way 1-shot}, \textit{5-way 5-shot}, \textit{5-way 20-shot}, and \textit{5-way 50-shot}. To ensure fair evaluation, all results in this paper are directly sourced from the original studies, and follow the setup in \cite{bscd-fsl}, with $C$-way $K$-shot data randomly sampled and averaged over 600 trials for robustness. As a subfield of FSL, many classical FSL methods are applied to CDFSL. Table~\ref{fsl_mtd} shows that meta-learning methods (e.g., MatchingNet, ProtoNet, RelationNet, MAML) perform worse in CDFSL due to domain gaps, while simple fine-tuning methods perform better as \textit{K} increases.
\begin{table}
\tiny
\centering
\vspace{-0.3cm}
\caption{CDFSL performance on the classical FSL approaches with ResNet10 backbone. $K$ means \textit{5-way K-shot}.}
\vspace{-0.3cm}
\setlength{\tabcolsep}{1.8mm}{
\begin{tabular}{clcccccccc}
\hline 
\textbf{\textit{K}} & \textbf{Methods} & \textbf{CropDiseases} & \textbf{EuroSAT} & \textbf{ISIC} & \textbf{ChestX} & \textbf{Plantae} & \textbf{Places} & \textbf{Cars} & \textbf{CUB}   \\ 
\hline
\multirow{5}*{1}  & Fine-tuning~\cite{bscd-fsl} & 61.56±0.90 & 49.34±0.85 & 30.80±0.59 &  21.88±0.38 & 33.53±0.36 & 50.87±0.48 & 29.32±0.34 &  41.98±0.41    \\ 
  & MatchingNet~\cite{miniimagenet} & 48.47±1.01 & 50.67±0.88 & 29.46±0.56 & 20.91±0.30 & 32.70 ± 0.60 & 49.86±0.79 & 30.77±0.47 & 35.89±0.51 \\
  &  RelationNet~\cite{relation} & 56.18±0.85 & 56.28±0.82 & 29.69±0.60 & 21.94±0.42 & 33.17±0.64 & 48.64±0.85 & 29.11±0.60 & 42.44±0.77 \\
  &  ProtoNet~\cite{proto} & 51.22±0.50 & 52.93±0.50 & 29.20±0.30 & 21.57±0.20 & - & - & - & - \\ 
  &  GNN~\cite{gnn} & \textbf{64.48±1.08} & \textbf{63.69±1.03} & \textbf{32.02±0.66} & \textbf{22.00±0.46} & \textbf{35.60±0.56} & \textbf{53.10±0.80} & \textbf{31.79±0.51} & \textbf{45.69±0.68}  \\ 
 \hline
\multirow{6}*{5}  & Fine-tuning~\cite{bscd-fsl} & \textbf{90.64±0.54} & 81.76±0.48 &  \textbf{49.68±0.36} & \textbf{26.09±0.96} & \textbf{47.40±0.36} & 66.47±0.41 & 38.91±0.38 &  58.75±0.36   \\ 
  & MatchingNet~\cite{miniimagenet} & 66.39±0.78 & 64.45±0.63 & 36.74±0.53 & 22.40±0.70 & 46.53±0.68 & 63.16±0.77 & 38.99±0.64 & 51.37±0.77   \\ 
  &  MAML~\cite{al21} & 78.05±0.68 & 71.70±0.72 & 40.13±0.58 & 23.48±0.96 & - & - & - & 47.20±1.10   \\ 
  &  RelationNet~\cite{relation} & 68.99±0.75 & 61.31±0.72 & 39.41±0.58 & 22.96±0.88 & 44.00±0.60 & 63.32±0.76 & 37.33±0.68 & 57.77±0.69  \\ 
  &  ProtoNet~\cite{proto} &  79.72±0.67 & 73.29±0.71 & 39.57±0.57 & 24.05±1.01 & - & - & - & \textbf{67.00±1.00}    \\ 
  &  GNN~\cite{gnn} & 87.96±0.67 & \textbf{83.64±0.77} & 43.94±0.67 & 25.27±0.46 & 52.53±0.59 & \textbf{70.84±0.65} & \textbf{44.28±0.63} & 62.25±0.65    \\ 
 \hline
\multirow{5}*{20}  & Fine-tuning~\cite{bscd-fsl} & \textbf{95.91±0.72} & \textbf{87.97±0.42} & \textbf{61.09±0.44} & \textbf{31.01±0.59} & - & - & - & - \\ 
  & MatchingNet~\cite{miniimagenet} & 76.38±0.67 & 77.10±0.57 &  45.72±0.53 & 23.61±0.86 & - & - & - & -  \\ 
  &  MAML~\cite{al21} & 89.75±0.42 & 81.95±0.55 & 52.36±0.57 & 27.53±0.43 & - & - & - & -    \\ 
  &  RelationNet~\cite{relation} & 80.45±0.64 & 74.43±0.66 & 41.77±0.49 &  26.63±0.92 & - & - & - & -  \\ 
  &  ProtoNet~\cite{proto}  &  88.15±0.51 & 82.27±0.57 & 49.50±0.55 & 28.21±1.15 & - & - & - & -    \\ 
 \hline
\multirow{4}*{50}  & Fine-tuning~\cite{bscd-fsl} & \textbf{97.48±0.56} & \textbf{92.00±0.56} & \textbf{67.20±0.59} & \textbf{36.79±0.53} & - & - & - & -  \\ 
  & MatchingNet~\cite{miniimagenet} & 58.53±0.73 & 54.44±0.67 &  54.58±0.65 & 22.12±0.88 & - & - & - & -   \\ 
  &  RelationNet~\cite{relation} & 85.08±0.53 & 74.91±0.58 & 49.32±0.51 & 28.45±1.20 & - & - & - & -   \\ 
  &  ProtoNet~\cite{proto} & 90.81±0.43 & 80.48±0.57 & 51.99±0.52 & 29.32±1.12 & - & - & - & -    \\
 \hline
\end{tabular}}
\vspace{-0.3cm}
\label{fsl_mtd}
\end{table}

Due to varying requirements (\eg, datasets, backbones) and configurations (\eg, training sets, modules) of current CDFSL methods, unified and fair comparisons are challenging. Table~\ref{repre_mtd} summarizes key details and performance of representative methods on FGCB and BSCD-FSL, highlighting the best (in blue and red) and sub-optimal (in light blue and light red) results for \textit{5-way 1-shot} and \textit{5-way 5-shot}. On BSCD-FSL, hybrid methods excel in near-domain scenarios, while $\Delta$-Adaptation methods perform best in distant-domain settings. $\mathcal{D}$-Extension methods show sub-optimal performance on BSCD-FSL but achieve the best results on FGCB, benefiting from information augmentation in fine-grained migration.

\begin{figure}
	\centering
 \vspace{-0.3cm}
 \includegraphics[width=\linewidth]{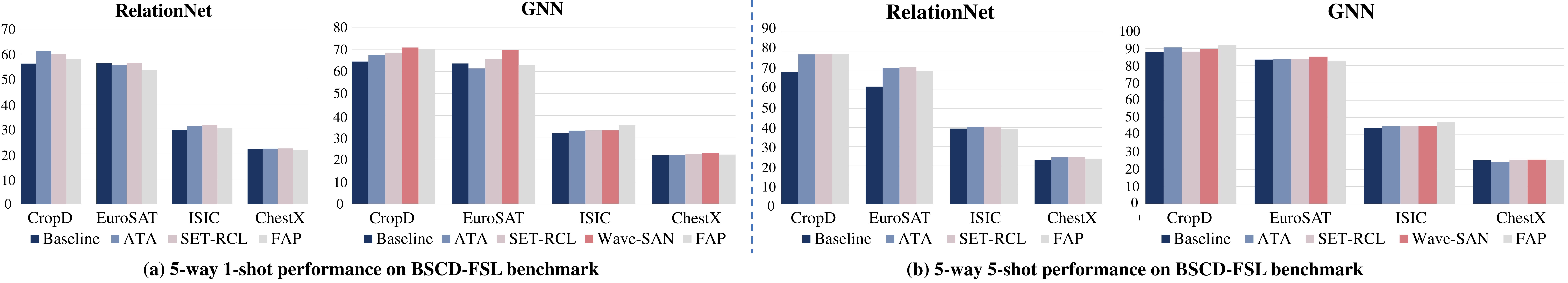}
 \vspace{-0.6cm}
	\caption{\textcolor{black}{We group by different FSL paradigms, \ie, MatchingNet, RelationNet, and GNN. The performance of different methods on the BSCD-FSL benchmark is presented in (a) \textit{5-way 1-shot} and (b) \textit{5-way 5-shot} tasks. All methods use ResNet10 as the backbone. `CropD' refers to the dataset `CropDiseases'.}} 
 \vspace{-0.4cm}
	\label{module-comp1}
\end{figure}

\begin{figure}
	\centering
 \vspace{-0.3cm}
        \includegraphics[width=\linewidth]{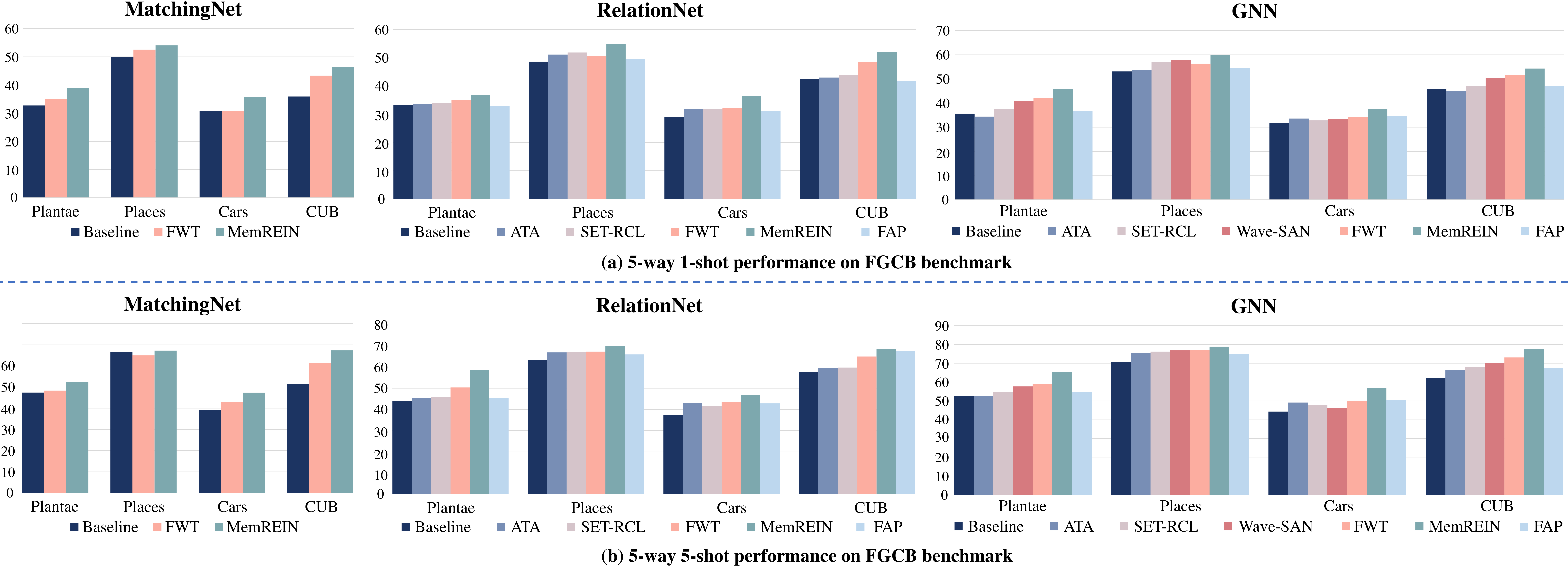}
 \vspace{-0.5cm}
	\caption{\textcolor{black}{The performance of various approaches on the FGCB benchmark is presented in (a) \textit{5-way 1-shot} and (b) \textit{5-way 5-shot} tasks. All methods use ResNet10 as the backbone. `CropD' refers to the dataset `CropDiseases.'}}
 \vspace{-0.4cm}
	\label{module-comp2}
\end{figure}

An emerging approach in CDFSL integrates plug-and-play modules into existing FSL paradigms like MatchingNet, RelationNet, and GNN. Figures~\ref{module-comp1} and~\ref{module-comp2} show five key findings: (1) GNN outperforms MatchingNet and RelationNet, excelling in CDFSL tasks; (2) performance drops as the domain distance increases (e.g., from CropDiseases to ChestX); (3) increasing samples (\textit{1-shot} to \textit{5-shot}) does not significantly improve distant-domain tasks; (4) Wave-SAN excels in near-domain tasks but shows diminishing returns with more data; and (5) MemREIN achieves the best results on fine-grained cross-domain datasets.

\begin{sidewaystable}
  \tiny
 \vspace{5.45in}
 \begin{center}
     \caption{The CDFSL performance of the proposed methods on BSCD-FSL and FGCB benchmarks. $K$ means \textit{5}-way \textit{K}-shot.}
     \vspace{-0.3cm}
\setlength{\tabcolsep}{0.5mm}{
\begin{tabular} {c|lcccccccccccc|m{4cm}}
\Xhline{1.2pt}
\textbf{Type} & \textbf{Methods} & \textbf{Venue} & \textbf{Train set} & \textbf{Backbone} & \textbf{$K$} & \textbf{CropDiseases} & \textbf{EuroSAT} & \textbf{ISIC} & \textbf{ChestX} & \textbf{Plantae} & \textbf{Places} & \textbf{Cars} & \textbf{CUB} & \textbf{Highlight}  \\ 
\Xhline{1.2pt}
\multirow{14}*{\rotatebox{90}{$\mathcal{D}$-Extension}} & \multirow{3}*{NSAE~\cite{boosting}} & \multirow{3}*{ICCV} & \multirow{3}*{\textit{mini}ImageNet} & \multirow{3}*{ResNet10} & 5 &  93.31±0.42 & 84.33±0.55 & \color{red!50}{\textbf{55.27±0.62}} & 27.30±0.42 & 62.15±0.77 & 73.17±0.72 & 58.30±0.75 & 71.92±0.77 & \multirow{3}{4.1cm}{The latent noise information from the source domain is utilized to capture broader variations of the feature distributions.}
\\
& & &  &  & 20 &  98.33±0.18 &  92.34±0.35 & 67.28±0.61 & 35.70±0.47 & 77.40±0.65 & 82.50±0.59 & 82.32±0.50 & 88.09±0.48 &    \\
& & & &  & 50 &  99.29±0.14 & 95.00±0.26 & 72.90±0.55 & 38.52±0.71 & 83.63±0.60 & 85.92±0.56 & - & 91.00±0.79 &   \\
\cline{2-15}
& \multirow{2}*{RDC~\cite{feature_reweight_6}} & \multirow{2}*{CVPR} & \multirow{2}*{\textit{mini}ImageNet} & \multirow{2}*{ResNet10} & 1 & \color{blue!50}{\textbf{86.33±0.50}} & 71.57±0.50 & 35.84±0.40 & 22.27±0.20 & 44.33±0.60 & \color{blue!50}{\textbf{61.50±0.60}} & 39.13±0.50 & 51.20±0.50 & \multirow{2}{4.1cm}{Minimising task-irrelevant features by constructing subspace} \\ 
&  &  &  &  & 5 & 93.55±0.30 & 84.67±0.30 & 49.06±0.30 & 25.48±0.20 & 60.63±0.40 & 74.65±0.40 & 53.75±0.50 & 67.77±0.40 &    \\
\cline{2-15}
& \multirow{2}*{StyleAdv~\cite{feature_reweight_0}} & \multirow{2}*{CVPR} & \multirow{2}*{\textit{mini}ImageNet} & \multirow{2}*{ResNet10} & 1 & 80.69±0.28 & 72.92±0.75 & 35.76±0.52 & 22.64±0.35 & 41.13±0.67 & 58.58±0.83 & 35.09±0.55 & 48.49±0.72 & \multirow{2}{4.1cm}{A meta Style Adversarial training method and a style adversarial attack method} \\ 
&  &  &  &  & 5 & \color{red!50}{\textbf{96.51±0.28}} & \color{red!50}{\textbf{91.64±0.43}} & 53.05±0.54 & 26.24±0.35 & 64.10±0.64 & 79.35±0.61 & 56.44±0.68 & 70.90±0.63 &    \\
\cline{2-15}
& \multirow{2}*{ISSNet~\cite{data_multi_3}} & \multirow{2}*{ICIP} & \multirow{2}*{\makecell[c]{\textit{mini}ImageNet \\ other 7 datasets}} & \multirow{2}*{ResNet10} & 1 & 73.40±0.86 & 64.50±0.88 & 36.06±0.69 & 23.23±0.42 & - & - & - & - & \multirow{2}{4.1cm}{Transferring styles across multiple sources to broaden the distribution of labeled sources} \\
&  &  &  &  & 5 & 94.10±0.41 & 83.64±0.55 & 51.82±0.67 &  \color{red!50}{\textbf{28.79±0.48}} &  -  & - & - & - &   \\
\cline{2-15}
& \multirow{2}*{STARTUP~\cite{st}} & \multirow{2}*{ICLR} & \multirow{2}*{\makecell[c]{\textit{mini}ImageNet \\ target data}} & \multirow{2}*{ResNet10} & 1 & 75.93±0.80 & 63.88±0.84 & 32.66±0.60 & 23.09±0.43 & - & - & - & - & \multirow{2}{4.1cm}{Self-training a source representation using unlabeled data from the target domain}  \\ 
&  &  &  &  & 5 & 93.02±0.45 &  82.29±0.60 & 47.22±0.61 & 26.94±0.44 & - & - & - & - &   \\
\cline{2-15}
& \multirow{2}*{TGDM~\cite{tgdm}} & \multirow{2}*{ACM MM} & \multirow{2}*{\textit{mini}ImageNet} & \multirow{2}*{ResNet10} & 1 & - & - & - & - & \color{blue!50}{\textbf{52.39±0.25}} & \color{blue}{\textbf{61.88±0.26}} & 50.70±0.24 & \color{blue!50}{\textbf{64.80±0.26}} & \multirow{2}{4.1cm}{A method generates an intermediate domain generation to facilitate the FSL task}  \\
&  &  &  &  & 5 & - & - & - & - & 71.78±0.22 & \color{red}{\textbf{81.62±0.19}} & 70.99±0.21 & \color{red}{\textbf{84.21±0.18}} &   \\
\cline{2-15}
& HVM~\cite{parameter_weight_2} & ICLR & \makecell[c]{\textit{mini}ImageNet} & ResNet10 & 5 & 87.65±0.35 & 74.88±0.45 & 42.05±0.34 & 27.15±0.45 & - & - & - & - & Introducing a hierarchical prototype model and an alternative method to solve domain gaps by using features at different semantic levels  \\
\Xhline{1.2pt}

\multirow{11}*{\rotatebox{90}{$\mathcal{H}$-Constraint}} & \multirow{3}*{SB-MTL~\cite{parameter_fix_1}} & \multirow{3}*{arXiv} & \multirow{3}*{\textit{mini}ImageNet} & \multirow{3}*{ResNet10} & 5 & 96.01±0.40 & 87.30±0.68 &  53.50±0.79 &  28.08±0.50 & - & - & - & - & \multirow{3}{4.1cm}{Leveraging a first-order MAML algorithm to identify optimal initializations and employing a score-based GNN for prediction}    \\
& &  &  &  & 20 &  99.61±0.09 &  96.53±0.28 & 70.31±0.72 & 37.70±0.57 & - & - & - & - &  \\
& &  &  &  & 50 & 99.85±0.06 & 98.37±0.18 & 78.41±0.66 & 43.04±0.66 & - & - & - & - &   \\
 \cline{2-15}
& \multirow{2}*{ReFine~\cite{parameter_weight_1}} & \multirow{2}*{ICMLW} & \multirow{2}*{\textit{mini}ImageNet} & \multirow{2}*{ResNet10} & 1 & 68.93±0.84 & 64.14±0.82 & 35.30±0.59 & 22.48±0.41 & - & - & - & - & \multirow{2}{4.1cm}{Randomizing the fitted parameters from the source domain before adapting to target data}  \\ 
&  &  &  &  & 5 & 90.75±0.49 & 82.36±0.57 & 51.68±0.63 & 26.76±0.42 & - & - & - & - &   \\
\cline{2-15}
& \multirow{2}*{ME-D2N~\cite{data_target}} & \multirow{2}*{ACM MM} & \multirow{2}*{\textit{mini}ImageNet} & \multirow{2}*{ResNet10} & 1 & - & - & - & - & \color{blue}{\textbf{52.89±0.83}} & 60.36±0.86 & 49.53±0.79 & \color{blue}{\textbf{65.05±0.83}} & \multirow{2}{4.1cm}{AME-D2N utilizes a multi-expert learning approach to create a model}  \\ 
&  &  &  &  & 5 & - & - & - & - & \color{red}{\textbf{72.87±0.67}} & \color{red!50}{\textbf{80.45±0.62}} & 69.17±0.68 & \color{red!50}{\textbf{83.17±0.56}} &   \\
\cline{2-15}
 & \multirow{2}*{MC-TS~\cite{wang2024cross}} & \multirow{2}*{EngA} & \multirow{2}*{\textit{mini}ImageNet} & \multirow{2}*{ResNet10} & 1 & 74.36±0.88 &  63.21±0.88 & 33.87±0.59 & \color{blue!50}{\textbf{23.61±0.42}} & - & - & - & - & \multirow{2}{4.1cm}{Match the large and small image crops, which are predicted by teacher and student network} \\
&  &  &  &  & 5 & 92.20±0.54 & 81.52±0.60 & 47.68±0.62 & 27.06±0.37& - &  - & - & - &    \\
\cline{2-15}
&  \multirow{2}*{LRP~\cite{feature_reweight_1}} & \multirow{2}*{ICPR} & \multirow{2}*{\textit{mini}ImageNet} & \multirow{2}*{ResNet10} & 1 & - & - & - & - & 34.80±0.37 &  50.59±0.46 & 29.65±0.33 & 42.44±0.41 & \multirow{2}{4.1cm}{A training strategy guided by explanations is developed to identify important features} \\  
& &  &  &  & 5 & - & - & - & - & 48.09±0.35 & 66.90±0.40 & 39.19±0.38 & 59.30±0.40 & \\
\Xhline{1.2pt} 

\multirow{20}*{\rotatebox{90}{$\Delta$-Adaptation}} 
&  \multirow{3}*{DARA~\cite{zhao2023dual}} & \multirow{3}*{TPAMI} & \multirow{3}*{\textit{mini}ImageNet} & \multirow{3}*{ResNet10} & 1 &  81.50±0.66 &  69.39±0.84 & \color{blue!50}{\textbf{38.49±0.66}} &  22.93±0.40 & 51.25±0.58 & 42.08±0.55 & \color{blue!50}{\textbf{52.70±0.83}} & 35.25±0.57 & \multirow{3}{4.1cm}{Focus on the fast adaptation capability of meta-learners by proposing an effective dual adaptive representation alignment approach} \\ 
&  &  &  &  & 5  &  96.23±0.34 & 87.67±0.54 & \color{red}{\textbf{57.54±0.68}} & 28.78±0.45 & \color{red!50}{\textbf{72.15±0.43}} & 65.40±1.95 & \color{red}{\textbf{77.51±0.65}} & 58.44±2.39 &   \\
&  &  &  &  & 20 &  99.21±0.11 & 94.40±0.27 & 68.43±0.54 & 36.20±0.43 & 83.12±0.54 & 79.54±0.64 & 89.60±0.43 & 81.38±0.59 &    \\
\cline{2-15}
&  \multirow{4}*{VDB~\cite{parameter_fix_6}} & \multirow{4}*{CVPRW} & \multirow{4}*{\textit{mini}ImageNet} & \multirow{4}*{ResNet10} & 1 &  71.98±0.82 &  63.60±0.87 & 35.32±0.65 &  22.99±0.44 & - & - & - & - & \multirow{4}{4.1cm}{Propose a source-free approach through ``Visual Domain Bridge'' concept to mitigate internal mismatches in BatchNorm during cross-domain settings} \\ 
&  &  &  &  & 5  &  90.77±0.49 & 82.06±0.63 & 48.72±0.65 & 26.62±0.45 &   & - & - & - &   \\
&  &  &  &  & 20 &  96.36±0.27 & 89.42±0.45 & 59.09±0.59 & 31.87±0.44 & - & - & - & - &    \\
&  &  &  &  & 50 & 97.89±0.19 & 92.24±0.35 & 64.02±0.58 & 35.55±0.45 & - & - & - & - &    \\
 \cline{2-15}
 & \multirow{2}*{MAP~\cite{parameter_select_2}} & \multirow{2}*{arXiv} & \multirow{2}*{\textit{mini}ImageNet} & \multirow{2}*{ResNet10} & 5 & 90.29±1.56 & 82.76±2.00 &  47.85±1.95 &  24.79±1.22 & 58.45±1.15 & 75.94±0.97 & 51.64±1.16 & 67.92±1.10 & \multirow{2}{4.1cm}{Selectively performs SOTA adaptation methods in sequence with modular adaptation method} \\  
 &  &  &  &  & 20 & 95.22±1.13 & 88.11±1.78 &  60.16±2.70 & 30.21±1.78 & - & - & - & - &   \\
 \cline{2-15}
& \multirow{3}*{TACDFSL~\cite{feature_reweight_4}} & \multirow{3}*{Symmetry} & \multirow{3}*{\textit{mini}ImageNet} & \multirow{3}*{WideResNet} & 5 & 93.42±0.55 & 85.19±0.67 & 45.39±0.67 & 25.32±0.48 & - & - & - & - & \multirow{3}{4.1cm}{Introducing the empirical marginal distribution measurement}  \\
&  &  &  &  & 20 & 95.49±0.39 & 87.87±0.49 & 53.15±0.59 & 29.17±0.52 & - & - & - & - &   \\
 &  &  &  &  & 50 & 95.88±0.35 & 89.07±0.43 &  56.68±0.58 & 31.75±0.51 & - & - & - & - &   \\
\cline{2-15}
& \multirow{2}*{IM-DCL~\cite{xu2024enhancing}} & \multirow{2}*{TIP} & \multirow{2}*{\makecell[c]{\textit{mini}ImageNet}} & \multirow{2}*{ResNet10} & 1 & 84.37±0.99 & \color{blue!50}{\textbf{77.14±0.71}} & 38.13±0.57 & \color{blue}{\textbf{23.98±0.79}} & - & - & - & - & \multirow{2}{4.1cm}{Enhancing information maximization with a distance-aware contrastive learning}  \\
&  &  &  &  & 5 & 95.73±0.38 & 89.47±0.42 & 52.74±0.69 & \color{red}{\textbf{28.93±0.41}} & - & - & - & - &   \\
\cline{2-15}
& \multirow{3}*{Confess~\cite{confess}} & \multirow{3}*{ICLR} & \multirow{3}*{\textit{mini}ImageNet} & \multirow{3}*{ResNet10} & 5 & 88.88±0.51 & 84.65±0.38 & 48.85±0.29 & 27.09±0.24 & - & - & - & - & \multirow{3}{4.1cm}{Investigating a contrastive learning and feature selection system to address domain gaps between base and novel categories} \\
&  &  &  &  & 20 & 95.34±0.48 & 90.40±0.24 & 60.10±0.33 & 33.57±0.31 & - & - & - & - &   \\
 &  &  &  &  & 50 & 97.56±0.43 & 92.66±0.36 & 65.34±0.45 & 39.02±0.12 & - & - & - & - &    \\
\cline{2-15}
& BL-ES~\cite{feature_reweight_2} & ICME & \makecell[c]{\textit{mini}ImageNet}  & ResNet18 & 5 & - & 79.78±0.83 & - & - & - & - & 50.07±0.84 & 69.63±0.88 & Proposing a bi-level episode strategy to train an inductive graph network  \\
\cline{2-15}
& \multirow{2}*{DSL~\cite{data_target_1}} & \multirow{2}*{ICLR} & \multirow{2}*{\makecell[c]{\textit{mini}ImageNet}} & \multirow{2}*{ResNet10} & 1 & - & - & - & - & 41.17±0.80 & 53.16±0.88 & 37.13±0.69 & 50.15±0.80 & \multirow{2}{4.1cm}{Incorporating the cross-domain scenario into the training stage by rapidly switching targets}  \\
&  &  &  &  & 5 & - & - & - & - & 62.10±0.75 & 74.10±0.72 & 58.53±0.73 & 73.57±0.65 &   \\
\Xhline{1.2pt} 

\multirow{14}*{\rotatebox{90}{Hybrid}} & \multirow{2}*{FDMixup~\cite{hybrid_1}} & \multirow{2}*{ACM MM} & \multirow{2}*{\textit{mini}ImageNet} & \multirow{2}*{ResNet10} & 1 & 66.23±1.03 & 62.97±1.01 & 32.48±0.64 & 22.26±0.45 & 37.89±0.58 & 53.57±0.75 & 31.14±0.51 & 46.38±0.68 & \multirow{2}{4.1cm}{Utilizing few labeled target data to guide the model learning} \\ 
&  &  &  &  & 5 & 87.27±0.69 & 80.48±0.79 & 44.28±0.66 & 24.52±0.44 & 54.62±0.66 & 73.42±0.65 & 41.30±0.58 & 64.71±0.68 &    \\
\cline{2-15}
 & \multirow{2}*{TL-SS~\cite{hybrid_3}} & \multirow{2}*{AAAI} & \multirow{2}*{\textit{mini}ImageNet} & \multirow{2}*{ResNet10} & 1 & - &  65.73 & - & - & - & 55.83 & 33.22 & 45.92 & \multirow{2}{4.1cm}{Introducing a domain-irrelevant self-supervised learning method} \\
&  &  &  &  & 5 & - & 79.36 & - & - & - &  76.33 & 49.82 & 69.16 &    \\
\cline{2-15}
& \multirow{2}*{FGNN~\cite{parameter_fix_7}} & \multirow{2}*{KBS} & \multirow{2}*{\textit{mini}ImageNet} & \multirow{2}*{ResNet10} & 1 & - & - & -  & - & 41.44±0.69 & 56.74±0.82 & 34.37±0.60 & 52.97±0.75 & \multirow{2}{4.1cm}{Investigating instance normalization and the restitution module to enhance performance} \\  
&  &  &  &  & 5 & - & - & - & - & 60.81±0.66 & 76.12±0.63 & 50.19±0.69 & 71.99±0.64 &   \\
\cline{2-15}
& \multirow{2}*{DDA~\cite{dynamic}} & \multirow{2}*{NIPS} & \multirow{2}*{\makecell[c]{\textit{mini}ImageNet \\ target data}} & \multirow{2}*{ResNet10} & 1 & 82.14±0.78 & 73.14±0.84 & 34.66±0.58 & 23.38±0.43 & - & - & - & - & \multirow{2}{4.1cm}{Propose a dynamic distillation-based approach to enhance utilize unlabeled target data} \\
&  &  &  &  & 5 & 95.54±0.38 & 89.07±0.47 & 49.36±0.59 & 28.31±0.46 & - & - & - & - &    \\
\cline{2-15}
& \multirow{2}*{CLDFD~\cite{cdfsl231}} & \multirow{2}*{ICLR} & \multirow{2}*{\textit{mini}ImageNet} & \multirow{2}*{ResNet10} & 1 &  \color{blue}{\textbf{90.48±0.72}} & \color{blue}{\textbf{82.52±0.76}} & \color{blue}{\textbf{39.70±0.69}} & 22.39±0.44 & - & - & - & - & \multirow{2}{4.1cm}{A cross-level knowledge distillation method and feature denoising operation}  \\ 
&  &  &  &  & 5 & \color{red}{\textbf{96.58±0.39}} & \color{red}{\textbf{92.89±0.34}} &  52.29±0.62 & 25.98±0.43 & - & - & - & - &   \\
\cline{2-15}
 & \multirow{2}*{ProD~\cite{ma2023prod}} & \multirow{2}*{CVPR} & \multirow{2}*{\makecell[c]{\textit{mini}ImageNet}} & \multirow{2}*{ResNet10 \& ViT} & 1 & - & - & - & - & 42.86±0.59 & 53.92±0.72 & 38.02±0.63 & 53.97±0.71 & \multirow{2}{4.1cm}{Using the prompts to disentangle the domain-general and -specific knowledge} \\
&  &  &  &  & 5 & - & - & - & - & 65.82±0.65 & 75.00±0.72 & 59.49±0.68 & 79.19±0.59 &    \\
 \cline{2-15} 
 & \multirow{2}*{GMeta-FDMixup~\cite{hybrid_7}} & \multirow{2}*{TIP} & \multirow{2}*{\makecell[c]{\textit{mini}ImageNet \\ Labeled target data}} & \multirow{2}*{ResNet10} & 1 & - & - & - & - & 51.28±0.36 & 59.39±0.37 & \color{blue}{\textbf{53.10±0.35}} & 63.85±0.37 & \multirow{2}{4.1cm}{Utilize extra labeled target data and a Meta-FDMixup network to solve CDFSL} \\
&  &  &  &  & 5 & - & - & - & - & 69.45±0.30 & 78.80±0.28 & \color{red!50}{\textbf{71.80±0.30}} & 80.48±0.27 &    \\
\Xhline{1.2pt}
\end{tabular}}
\label{repre_mtd}
 \end{center}
 \end{sidewaystable}

\vspace{-0.2cm}
\subsubsection{Evaluation for $\mathcal{D}$-Extension Approaches}
Table~\ref{repre_mtd} highlights a noticeable trend where performance decreases as the distance between the target and source domains increases. For instance, the results of \cite{boosting} show a drop from 93.31\% on CropDiseases to 27.30\% on ChestX (\textit{5-way 5-shot}). The sub-optimal performances of \cite{feature_reweight_6} and \cite{feature_reweight_0} on CropDiseases (96.51\% for \cite{feature_reweight_0} on \textit{5-way 5-shot}) and EuroSAT (91.54\% for \cite{feature_reweight_0} on \textit{5-way 5-shot}) reveal that specific guidance in information extension can further improve performance in near-domain scenarios. Additionally, the optimal results on FGCB for \cite{tgdm} also indicate the positive effect of information augmentation on CDFSL. Furthermore, the results of \cite{st} on BSCD-FSL demonstrate that incorporating target domain data into the training process can improve performance in the target domain. However, this approach works better for near-domain transfer than for distant-domain transfer. For example, \cite{st} showed a 2.38\% improvement on CropDiseases and a 0.53\% improvement on EuroSAT, but a 2.46\% drop on ISIC and only a 0.85\% improvement on ChestX when compared to the classic fine-tuning method.

$\mathcal{D}$-Extension approaches for CDFSL are relatively simple in concept as they rely on adding supplementary information to enhance the model's generalization. However, their effectiveness is highly dependent on the choice of information used in the training process. If the additional information included in training greatly diverge from the target domain or the selected target domain samples are not representative, this fails to improve CDFSL performance.

\vspace{-0.2cm}
\subsubsection{Evaluation for $\mathcal{H}$-Constraint Approaches}
From the data presented in Table~\ref{repre_mtd}, it appears that the performance of $\mathcal{H}$-Constraint methods is generally subpar in comparison to the other two method types on BSCD-FSL. Using ResNet10 as the backbone, the results of ~\cite{parameter_fix_1} on BSCD-FSL (\textit{5-way 5-shot}) demonstrate this trend, with scores of 96.01\% (CropDiseases), 87.30\% (EuroSAT), 53.50\% (ISIC), and 28.08\% (ChestX). The results of other methods within this category are even lower. However,~\cite{wang2024cross} achieves sub-optimal results of 23.61\% on ChestX in the 5-way 1-shot task. Moreover,~\cite{data_target} performs best on FGCB, achieving optimal results of 52.89\% (Plantae), 65.05\% (CUB) in the 5-way 1-shot task, and 72.87\% (Plantae) in the 5-way 5-shot task. These findings demonstrate the potential of knowledge distillation in distant-domain scenarios.

Our analysis of Table~\ref{repre_mtd} shows that $\mathcal{H}$-Constraint methods generally underperform compared to other categories, likely due to their reliance on locally adjusting network parameters with limited additional parameters, which restricts adaptation to new domains~\cite{paszke2019pytorch}. While knowledge distillation methods achieve competitive results, $\mathcal{H}$-Constraint methods face challenges in solving the two-stage empirical risk minimization problem. Further exploration of knowledge distillation techniques is needed to enhance the performance of $\mathcal{H}$-Constraint methods in CDFSL.

\vspace{-0.2cm}
\subsubsection{Evaluation for $\Delta$-Adaptation Approaches}
The performance of recent $\Delta$-Adaptation approaches has a significant advantage on distant-domain tasks compared to other methods. This can be exemplified by comparing two representative approaches: \cite{zhao2023dual} achieved sub-optimal performance on ISIC in the 5-way 1-shot task (38.49\%) and optimal performance in the 5-way 5-shot task (57.54\%). Meanwhile, \cite{xu2024enhancing} obtained the best results for ChestX in both the 1-shot (23.98\%) and 5-shot (28.93\%) tasks. In addition to BSCD-FSL, $\Delta$-Adaptation approaches also show overwhelming performance advantages in fine-grained migration scenarios. For instance,~\cite{zhao2023dual} achieved competitive results on Plantae (72.15\% in the 5-way 5-shot task) and Cars (77.51\% in the 5-way 5-shot task). These observations suggest that $\Delta$-Adaptation methods are more effective than other approaches in addressing distant-domain tasks. However, not all methods in the $\Delta$-Adaptation category perform better, indicating that the design and application of $\Delta$-Adaptation strategies have a significant impact on performance.
In summary, compared to other strategies, $\Delta$-Adaptation is the most effective for addressing distant-domain tasks. Furthermore, $\Delta$-Adaptation is also effective in handling few-shot tasks in fine-grained migration scenarios.

\vspace{-0.2cm}
\subsubsection{Evaluation for Hybrid Approaches}
Hybrid methods demonstrate optimal performance on near-domain tasks. For instance, \cite{cdfsl231} achieves the best results on CropDiseases (90.48\% in the 1-shot task and 96.58\% in the 5-shot task) and EuroSAT (82.52\% in the 1-shot task and 92.89\% in the 5-shot task). However, this advantage diminishes as the domain gap widens. It is important to recognize that combining strategies from different categories introduces a degree of risk, as negative interactions between strategies can occur \cite{azevedo2024hybrid}. Thus, when selecting hybrid strategies, researchers must carefully consider the compatibility of different methods, especially regarding the precision required for alignment. Ultimately, the choice of approach depends on the specific task, available data, and the level of generalization and flexibility required by the model.

\vspace{-0.2cm}
\subsubsection{Evaluation on Meta-Dataset}
The techniques tested on Meta-Dataset~\cite{meta-dataset} used non-episodic training, with results summarized in Table~\ref{meta_per}. In the single-source setup (ImageNet as the source domain), \cite{feature_select_2} achieved the best performance on five target datasets, while \cite{parameter_weight_4, hybrid_2, yang2024leveraging} excelled on the other five. Deeper backbones like ResNet34~\cite{parameter_weight_4} outperformed shallower ones like ResNet18. In the multiple-source setup (first eight datasets as source domains), \cite{wang2023mmt} achieved the highest performance on three datasets, benefiting from effective feature transformation and multi-source combination. However, comparisons for \cite{parameter_weight_4} showed that while multiple datasets improved performance on seen datasets, gains on unseen datasets were modest, highlighting that adding domains without careful consideration may limit improvements. Overall, CDFSL technologies significantly outperform traditional FSL methods, effectively addressing $FG$ and $Art$ CDFSL challenges.
\begin{table}
\tiny
\centering
\vspace{-0.3cm}
\caption{The CDFSL performance of approaches on Meta-Dataset. Gray background means the results on the seen data set (source data set).}
\vspace{-0.3cm}
\setlength{\tabcolsep}{0.15mm}{
\begin{tabular}{lccccccc|cccccc}
\hline
 & \multicolumn{7}{c|}{\textbf{Single source}} & \multicolumn{6}{c}{\textbf{Multiple sources}}    \\ 
\cline{2-14}
 & ProtoNet~\cite{proto} & Pro-M~\cite{meta-dataset} & SUR~\cite{feature_select_2} & \multicolumn{2}{c}{TPA~\cite{parameter_weight_4}} & tri-M~\cite{hybrid_2} & ProLAD~\cite{yang2024leveraging} & RMFS~\cite{feature_select_1} & TPA~\cite{parameter_weight_4} & URL~\cite{hybrid_4} & TSA~\cite{sreenivas2023similar} & TA2-Net~\cite{guo2023task} & MMT~\cite{wang2023mmt}   \\ 
\hline
Backbone & ResNet18 & ResNet18 & ResNet18 & ResNet18 & ResNet34 & ResNet18 & ResNet18 & ResNet18 & ResNet18 & ResNet18 & ResNet18 & ResNet18 & ResNet18    \\ 
\hline
ImageNet & \cellcolor{gray!40}{44.5$\pm$1.1} & \cellcolor{gray!40}{47.9$\pm$1.1} & \cellcolor{gray!40}{57.2$\pm$1.1} & \cellcolor{gray!40}{\textbf{59.5$\pm$1.1}} & \cellcolor{gray!40}{63.7$\pm$1.0} & \cellcolor{gray!40}{58.6$\pm$1.0} & \cellcolor{gray!40}{57.2$\pm$1.1} & \cellcolor{gray!40}{\textbf{63.1$\pm$0.8}} & \cellcolor{gray!40}{59.5$\pm$1.0} & \cellcolor{gray!40}{58.8$\pm$1.1} & \cellcolor{gray!40}{59.5$\pm$1.0} & \cellcolor{gray!40}{59.6$\pm$1.0} & \cellcolor{gray!40}{59.6$\pm$1.1}   \\ 
\cline{2-8}
Omniglot & 79.6$\pm$1.1 & 82.9$\pm$0.9 & \textbf{93.2$\pm$0.8} & 78.2$\pm$1.2 & 82.6$\pm$1.1 & 92.0$\pm$0.6 & 84.1$\pm$1.2 & \cellcolor{gray!40}{\textbf{97.7$\pm$0.5}} & \cellcolor{gray!40}{94.9$\pm$0.4} & \cellcolor{gray!40}{94.5$\pm$0.4} & \cellcolor{gray!40}{94.9$\pm$0.4} & \cellcolor{gray!40}{95.5$\pm$0.4} & \cellcolor{gray!40}{94.4$\pm$0.4}   \\ 

Aircraft & 71.1$\pm$0.9 & 74.2$\pm$0.8 & \textbf{90.1$\pm$0.8} & 72.2$\pm$1.0 & 80.1$\pm$1.0 & 82.8$\pm$0.7 & 76.1$\pm$1.2 & \cellcolor{gray!40}{65.1$\pm$0.3} & \cellcolor{gray!40}{89.9$\pm$0.4} & \cellcolor{gray!40}{89.4$\pm$0.4} & \cellcolor{gray!40}{89.9$\pm$0.4} & \cellcolor{gray!40}{90.5$\pm$0.4} & \cellcolor{gray!40}{\textbf{91.9$\pm$0.5}}   \\ 

Birds & 67.0$\pm$1.0 & 70.0$\pm$1.0 & \textbf{82.3$\pm$0.8} & 74.9$\pm$ 0.9 & 83.4$\pm$0.8 & 75.3$\pm$0.8 & 75.5$\pm$1.0 & \cellcolor{gray!40}{\textbf{84.1$\pm$0.6}} & \cellcolor{gray!40}{81.1$\pm$0.8} & \cellcolor{gray!40}{80.7$\pm$0.8} & \cellcolor{gray!40}{81.1$\pm$0.8} & \cellcolor{gray!40}{81.4$\pm$0.8} & \cellcolor{gray!40}{82.7$\pm$0.5}   \\ 

Textures & 65.2$\pm$0.8 & 67.9$\pm$0.8 & 73.5$\pm$0.7 & \textbf{77.3$\pm$0.7} & 79.6$\pm$0.7 & 71.2$\pm$0.8 & 77.7$\pm$0.8 & \cellcolor{gray!40}{67.5$\pm$0.9} & \cellcolor{gray!40}{77.5$\pm$0.7} & \cellcolor{gray!40}{77.2$\pm$0.7} & \cellcolor{gray!40}{77.5$\pm$0.7} & \cellcolor{gray!40}{77.4$\pm$0.7} & \cellcolor{gray!40}{\textbf{78.2$\pm$0.9}}   \\ 

Quick Draw & 65.9$\pm$0.9 & 66.6$\pm$0.9 & \textbf{81.9$\pm$1.0} & 67.6$\pm$0.9 & 71.0$\pm$0.8 & 77.3$\pm$0.7 & 70.6$\pm$0.1 & \cellcolor{gray!40}{\textbf{86.2$\pm$0.5}} & \cellcolor{gray!40}{81.7$\pm$0.6} & \cellcolor{gray!40}{82.5$\pm$0.6} & \cellcolor{gray!40}{81.7$\pm$0.6} & \cellcolor{gray!40}{82.5$\pm$0.6} & \cellcolor{gray!40}{83.1$\pm$0.2}   \\ 

Fungi & 40.3$\pm$1.1 & 42.0$\pm$1.1 & \textbf{67.9$\pm$0.9} & 44.7$\pm$1.0 & 51.4$\pm$1.2 & 48.5$\pm$1.0 & 46.8$\pm$1.2 & \cellcolor{gray!40}{62.5$\pm$0.6} & \cellcolor{gray!40}{66.3$\pm$0.8} & \cellcolor{gray!40}{\textbf{68.1$\pm$0.9}} & \cellcolor{gray!40}{66.3$\pm$0.8} & \cellcolor{gray!40}{66.3$\pm$0.9} & \cellcolor{gray!40}{66.2$\pm$0.6}   \\ 

VGG Flower & 86.9$\pm$0.7 & 88.5$\pm$1.0 & 88.4$\pm$0.9 & 90.9$\pm$0.6 & 94.0$\pm$0.5 & 90.5$\pm$0.5 & \textbf{92.9$\pm$0.6} & \cellcolor{gray!40}{86.3$\pm$0.3} & \cellcolor{gray!40}{92.2$\pm$0.5} & \cellcolor{gray!40}{92.0$\pm$0.5} & \cellcolor{gray!40}{92.2$\pm$0.5} & \cellcolor{gray!40}{\textbf{92.6$\pm$0.4}} & \cellcolor{gray!40}{90.7$\pm$0.4}   \\ 
\cline{9-14}
Traffic Sign & 46.5$\pm$1.0 & 34.2$\pm$1.3 & 67.4$\pm$0.8 & 82.5$\pm$0.8 & 81.7$\pm$0.9 & 78.0$\pm$0.6 & \textbf{89.4$\pm$0.9} & 73.7$\pm$0.4 & 82.8$\pm$1.0 & 63.3$\pm$1.2 & 82.8$\pm$1.0 & \textbf{87.4$\pm$0.8} & 85.1$\pm$0.2   \\ 

MSCOCO & 39.9$\pm$1.1 & 24.1$\pm$1.1 & 51.3$\pm$1.0 & \textbf{59.0$\pm$1.0} & 61.7$\pm$0.9 & 52.8$\pm$1.1 & 55.4$\pm$1.1 & 56.2$\pm$0.7 & 57.6$\pm$1.0 & 57.3$\pm$1.0 & 57.6$\pm$1.0 & 57.9$\pm$0.9 & \textbf{58.9$\pm$0.5}   \\ 

MNIST & - & - & 90.8$\pm$0.5 & 93.9$\pm$0.6 & 94.6$\pm$0.5 & \textbf{96.2$\pm$0.3} & 95.8$\pm$0.5 & - & 96.7$\pm$0.4 & 94.7$\pm$0.4 & 96.7$\pm$0.4 & 97.0$\pm$0.4 & \textbf{97.3$\pm$0.3}   \\ 

CIFAR 10 & - & - & 66.6$\pm$0.8 & \textbf{82.1$\pm$0.7} & 86.0$\pm$0.6 & 75.4$\pm$0.8 & 79.7$\pm$0.8 & - & \textbf{82.9$\pm$0.7} & 74.2$\pm$0.8 & 82.9$\pm$0.7 & 82.1$\pm$0.8 & 82.0$\pm$0.6   \\ 

CIFAR 100 & - & - & 58.3$\pm$1.0 & \textbf{70.7$\pm$0.9} & 78.3$\pm$0.8 & 62.0$\pm$1.0 & 70.3$\pm$1.0 & - & 70.4$\pm$0.9 & 63.6$\pm$1.0 & 70.4$\pm$0.9 & 70.9$\pm$0.9 & \textbf{71.5$\pm$1.0}   \\ 
\hline
\end{tabular}}
\vspace{-0.3cm}
\label{meta_per}
\end{table}

\vspace{-0.2cm}
\section{Future work} \label{future}
Despite the significant progress made in CDFSL, it still presents unique challenges that require further investigation. Therefore, we outline several promising future research directions, focusing on problem setups, applications, and theoretical developments within the field.

\vspace{-0.2cm}
\subsection{Problem Setups}
\textit{Active Learning based CDFSL.}
In Section~\ref{challenge}, we discussed the challenge of CDFSL, primarily due to domain gap and task shifts, which is especially problematic when the domains are vastly different and target domain data is scarce. Addressing this challenge requires expanding and efficiently utilizing shared information between these domains. Active learning (AL)~\cite{10167762}, which identifies the most informative samples for labeling, has gained attention in both domain adaptation~\cite{ac_da,ac_da1} and few-shot learning~\cite{ac_fsl,ac_fsl1}. For example, \cite{ac_da} improves target domain recognition by prioritizing samples with high uncertainty and diversity. Similarly, \cite{ac_fsl} integrates FSL and AL into FASL, an iterative platform for efficient text classification. Given the ability of AI to select the most informative samples, it is well-suited to enhance both cross-domain and cross-task learning in CDFSL, making it a promising area for future research.

\textit{Source-free CDFSL.} 
Pre-trained models, such as CLIP, pose unique challenges in the CDFSL context due to the inaccessibility of their source data and pre-training strategies \cite{xu2024enhancing, zhuo2024prompt, xu2024step}. Existing CDFSL methods, which focus on improving model transferability by leveraging source data and optimizing pre-training strategies, are not well-suited to these conditions. Moreover, with growing concerns about data privacy, accessing source data in real-world scenarios may infringe on intellectual property rights. Therefore, it is imperative to develop a CDFSL approach that operates independently of source data and pre-training details. This necessitates the exploration of source-free cross-domain few-shot learning (SF-CDFSL) setting tailored to address these constraints effectively. In SF-CDFSL, the algorithms improve FSL performance on the target domain using only a pre-trained source model, without access to the source data. While this approach is well-established in domain adaptation~\cite{sfda1,sfda2,sfda3}, it remains in its early stages for CDFSL. Notable efforts include ~\cite{parameter_fix_6}, which adjusts batch normalization to align source and target distributions,~\cite{xu2024enhancing}, which uses information maximization for implicit alignment, and~\cite{zhuo2024prompt}, which uses limited target domain data to enhance diversity and applies VPT to adapt large-scale models. Given the unique challenges of CDFSL, further research in source-free settings is essential for safeguarding data privacy and usage of pre-trained models.

\textit{Prompt Tuning based CDFSL.} 
Fine-tuning is a common method for adapting pre-trained models in CDFSL, but it becomes costly as model sizes grow. To address this, prompt learning has emerged in natural language processing (NLP) \cite{ptnlp}. Prompt tuning (PT)~\cite{dapt1,zhuo2024prompt} uses minimal prompts to guide the model in understanding the target task, rather than fine-tuning all model parameters. This approach significantly reduces computational cost and storage requirements while maintaining competitive performance on downstream tasks. PT has proven effective in FSL tasks in NLP~\cite{ptnlpfsl1,ptnlpfsl2} and has shown promise in solving domain adaptation problems~\cite{dapt1} in computer vision. Recent works~\cite{zhuo2024prompt, xu2024step} have attempted to apply PT to address the CDFSL problem, achieving competitive performance. However, its effectiveness heavily depends on the prompt design, often requiring careful crafting or optimization to achieve optimal results. In summary, exploring the potential of PT in CDFSL represents a promising avenue for future research.

\textit{Incremental CDFSL.}
Current CDFSL methodologies focus on addressing FSL tasks in the target domain but often suffer from catastrophic forgetting, where performance on the source domain declines. A robust model should retain knowledge from both domains and tasks. Thus, addressing catastrophic forgetting in CDFSL is a critical challenge. Recent advances in incremental and continuous learning have been applied in FSL to tackle task-incremental issues~\cite{cdfsil_1,cdfsil_2}. For example, \cite{cdfsil_1} stabilizes network topology to minimize forgetting of previous classes, while \cite{cdfsil_2} focuses on updating classifiers incrementally, preventing the erasure of knowledge in the feature extractor. Building on these techniques, exploring domain-incremental learning in CDFSL is crucial. The goal is to extend the model to accommodate new domains and tasks while maintaining performance on prior ones.

\vspace{-0.2cm}
\subsection{Applications}

\textit{Rare Cancer Detection.}  
Cancer is a serious disease that demands early detection, and detecting rare cancers is particularly challenging due to the limited availability of data. While FSL has been employed for rare cancer detection~\cite{rcd1}, gathering sufficient auxiliary data from the same distribution as the target data can be challenging. This makes CDFSL an ideal solution, as it allows for the use of auxiliary data from different domains.

\textit{Intelligent Fault Diagnosis.}  
Intelligent fault diagnosis~\cite{app-diagnosis} involves detecting machine faults early using diagnostic methods. However, creating an ideal training dataset for diagnostic models is challenging. To address this challenge,~\cite{app-dia2} incorporated data from other domains and applied FSL algorithms. This makes intelligent fault diagnosis a promising application for CDFSL, allowing models to leverage auxiliary data and improve diagnostic accuracy in the face of limited training samples.

\textit{Solving Algorithmic Bias.}  
AI algorithms depend on training data to address many real-world problems, but inherent biases in the data can be learned and amplified by these algorithms. For example, when certain groups are underrepresented in a dataset, the algorithm may make poor predictions for those groups, which can lead to algorithmic bias~\cite{fairness}. This presents a critical ethical challenge in artificial intelligence. Ideally, AI systems should mitigate bias rather than exacerbate it. CDFSL offers a potential solution by focusing on reducing bias in datasets and generalizing across domains and tasks, effectively addressing domain and task shifts. Furthermore, CDFSL can mitigate performance loss caused by limited data from underrepresented groups, contributing to fairer outcomes.

\vspace{-0.3cm}
\subsection{Theories}
\vspace{-0.1cm}
\textit{Invariant Risk Minimization (IRM).}  
Machine learning systems often capture all correlations in the training data, including spurious correlations arising from data biases. To ensure generalization to new environments, eliminating such spurious correlations is essential. Invariant Risk Minimization (IRM)~\cite{irm} is a learning paradigm that addresses this by estimating nonlinear, invariant, causal predictors across multiple training environments, reducing a system’s dependence on data biases. In CDFSL, spurious correlations learned in the source domain should be discarded when adapting to tasks in the target domain. Thus, despite being in its early stages, IRM shows significant promise for CDFSL due to its emphasis on domain and task migration.

\textit{Multiple Source Domain Organization.}  
While some CDFSL approaches leverage multiple source domains to enhance FSL performance on the target domain, theoretical research on effectively organizing these source domains remains limited. Questions regarding how to select and optimize source domains for maximum FSL performance remain underexplored. Advancing theoretical work in this area could significantly enhance multi-source domain applications in CDFSL. A promising reference is provided by~\cite{mul_theory}, which offers a theoretical foundation for organizing multi-source domains. This could pave the way for more effective and rational strategies in multi-source domain CDFSL.

\textit{Domain Generalization.}  
The ultimate goal of CDFSL is to generalize not only to specific domains but to all domains. Theoretical research on domain generalization~\cite{generalization_theory} is crucial to achieving this. Leveraging this research can enable CDFSL to evolve into a few-shot domain generalization problem, allowing models to generalize across a wide range of domains, thereby enhancing their robustness and adaptability.
\vspace{-0.3cm}
\section{Conclusion} \label{conclusion}
Cross-domain few-shot learning (CDFSL) significantly enhances the capabilities of few-shot learning (FSL) by utilizing diverse domain samples to improve performance on target domains, thereby addressing the critical challenge of limited labeled data. This survey provides a thorough review of CDFSL, starting from foundational definitions and advancing to its formalization, while distinguishing it from related fields such as domain adaptation and multi-task learning. A primary challenge within CDFSL is the unreliable two-stage empirical risk minimization, which complicates the learning of effective shared features and hinders adaptability to new tasks. We categorize existing strategies into four main approaches: $\mathcal{D}$-Extension, $\mathcal{H}$-Constraint, and $\Delta$-Adaptation, along with hybrid methods, each presenting unique strengths and weaknesses that influence their effectiveness in different scenarios. Looking forward, research should focus on specific areas such as integrating active learning techniques to enhance shared knowledge acquisition, exploring source-free CDFSL approaches to safeguard data privacy, and utilizing prompt learning to streamline model adaptations. By addressing these challenges and exploring these avenues, CDFSL can significantly impact practical applications across various domains, paving the way for robust and adaptable machine learning solutions.

\bibliographystyle{ACM-Reference-Format}
\bibliography{sample-acmsmall}

\end{document}